\documentclass[times,twocolumn,final,longtitle]{elsarticle}
\usepackage{framed,multirow}
\usepackage{adjustbox}
\usepackage{amssymb}
\usepackage{latexsym}
\usepackage{url}
\usepackage[hidelinks]{hyperref}
\hypersetup{
  colorlinks   = true, 
  urlcolor     = blue, 
  linkcolor    = blue, 
  citecolor   = red 
}
\usepackage[utf8]{inputenc}
\usepackage[T1]{fontenc}
\usepackage{url}
\usepackage[hidelinks]{hyperref}
\hypersetup{
  colorlinks   = true, 
  urlcolor     = blue, 
  linkcolor    = blue, 
  citecolor   = red 
}
\usepackage{multirow}
\usepackage{multicol}
\usepackage{adjustbox}
\usepackage{lscape}
\usepackage{graphicx}			
\usepackage{subcaption}
\usepackage{longtable}
\usepackage{ltablex,booktabs}
\usepackage{makecell}
\usepackage{amsmath}			
\usepackage{amssymb}			
\usepackage{amsfonts}			
\usepackage{adjustbox}
\usepackage{lscape}
\usepackage{colortbl}
\usepackage{mathtools}
\usepackage[table]{xcolor}    

\DeclareMathOperator*{\argmin}{arg\,min} 
\DeclarePairedDelimiter{\norm}{\lVert}{\rVert}

\usepackage{newfloat}
\DeclareFloatingEnvironment[name={Supplementary Figure},fileext=lof]{suppfigure}
\DeclareFloatingEnvironment[name={Supplementary Table},fileext=lof]{suppTable}
\usepackage{lscape}
\usepackage{fltpage}
\usepackage{soul,color}
\usepackage{rotating}
\usepackage{commath}

\usepackage{multirow}

\journal{}
\begin{document}
\begin{frontmatter}
\title{An objective comparison of methods for augmented reality in laparoscopic liver resection by preoperative-to-intraoperative image fusion}
\author[1,14]{Sharib Ali}\corref{corr1}\cortext[corr1]{Corresponding author}
\ead{s.s.ali@leeds.ac.uk}
\author[2,14]{Yamid Espinel}
\author[5]{Yueming Jin}
\author[7]{Peng Liu}
\author[7]{Bianca Güttner}
\author[8]{Xukun Zhang}
\author[8]{Lihua Zhang}
\author[9]{Tom Dowrick} 
\author[9]{Matthew J. Clarkson} 
\author[10]{Shiting Xiao}
\author[11]{Yifan Wu}
\author[12]{Yijun Yang}
\author[12]{Lei Zhu}
\author[13]{Dai Sun}
\author[13]{Lan Li}
\author[7]{Micha Pfeiffer} 
\author[4]{Shahid Farid} 
\author[6]{Lena Maier-Hein}
\author[2,3]{Emmanuel Buc}
\author[2,3]{Adrien Bartoli}
\address[1]{School of Computing, Faculty of Engineering and Physical Sciences, University of Leeds, Leeds, LS2 9JT, United Kingdom}
\address[2]{Centre Hospitalier Universitaire de Clermont-Ferrand, Clermont-Ferrand, France}
\address[3]{Endoscopy and Computer Vision Group, Université Clermont Auvergne, Clermont-Ferrand, France}
\address[4]{Department of HPB and Transplant Surgery, St. James’s University Hospital, Leeds, United Kingdom}
\address[5]{Department of Electrical and Computer Engineering 
National University of Singapore (NUS), 119276, Singapore}
\address[6]{German Cancer Research Center (DKFZ), Heidelberg, Germany}
\address[7]{Department of Translational Surgical Oncology, National Center for Tumor Diseases (NCT/UCC Dresden), Fetscherstraße 74, 01307 Dresden}
\address[8]{Academy for Engineering and Technology, Fudan University, Shanghai, China}
\address[9]{Wellcome EPSRC Centre for Interventional
and Surgical Sciences, UCL, London, UK}
\address[10]{Department of Computer and Information Science, University of Pennsylvania, PA 19104, Philadelphia, US}
\address[11]{Department of Bioengineering, University of Pennsylvania, PA 19104, Philadelphia, US}
\address[12]{Hong Kong University of Science and Technology (Guangzhou), 511455, Guangzhou, China} 
\address[13]{Suzhou Institute for Advanced Research, Center for Medical Imaging, Robotics, Analytic Computing \& Learning (MIRACLE), University of Science and Technology of China, Suzhou, China} 
\address[14]{these authors contributed equally to this work}
\begin{abstract}
Augmented reality for {laparoscopic liver resection} is a {visualisation mode} that {allows} a surgeon to localise tumours and vessels embedded within {the} liver by projecting them on top of {a} laparoscopic image. During this process, preoperative 3D models extracted from CT or MRI data are registered to the intraoperative laparoscopic images. 
In terms of 3D-2D fusion, most of the algorithms make use of anatomical landmarks to guide registration. These landmarks include the liver's inferior ridge, the falciform ligament, and the occluding contours. They are usually marked by hand in both the laparoscopic image and the 3D model, which is time consuming and may contain errors if done by a non-experienced user. Therefore, there is a need {to automate} this process so that augmented reality can be used effectively in the operating room. {We} present the {Preoperative-to-Intraoperative} Laparoscopic Fusion challenge (P2ILF), held during the Medical Imaging and Computer Assisted Interventions (MICCAI 2022) conference, which {investigates} the possibilities of detecting these landmarks automatically and using them in registration. The challenge was divided into two tasks: 1) A 2D and 3D landmark detection task, and 2) a 3D-2D registration task. {The teams} were provided with a set of training data consisting of 167 laparoscopic images and 9 preoperative 3D models from 9 patients, with the corresponding 2D and 3D landmark annotations. A total of 6 teams from {4} countries participated in the challenge, whose results were assessed for each task independently. All the teams proposed deep learning-based methods for the 2D and 3D landmark segmentation tasks, and differentiable rendering-based methods for the registration task. The proposed methods were evaluated on a test set of 16 images and two preoperative 3D models from two patients. Based on the experimental outcomes, we propose three key {hypotheses} that determine {current} limitations and future directions for research in this domain.
\end{abstract}
\end{frontmatter}
%
\section*{Introduction}
\label{sec:introduction}
\begin{figure*}
    \centering
    \includegraphics[width=\textwidth]{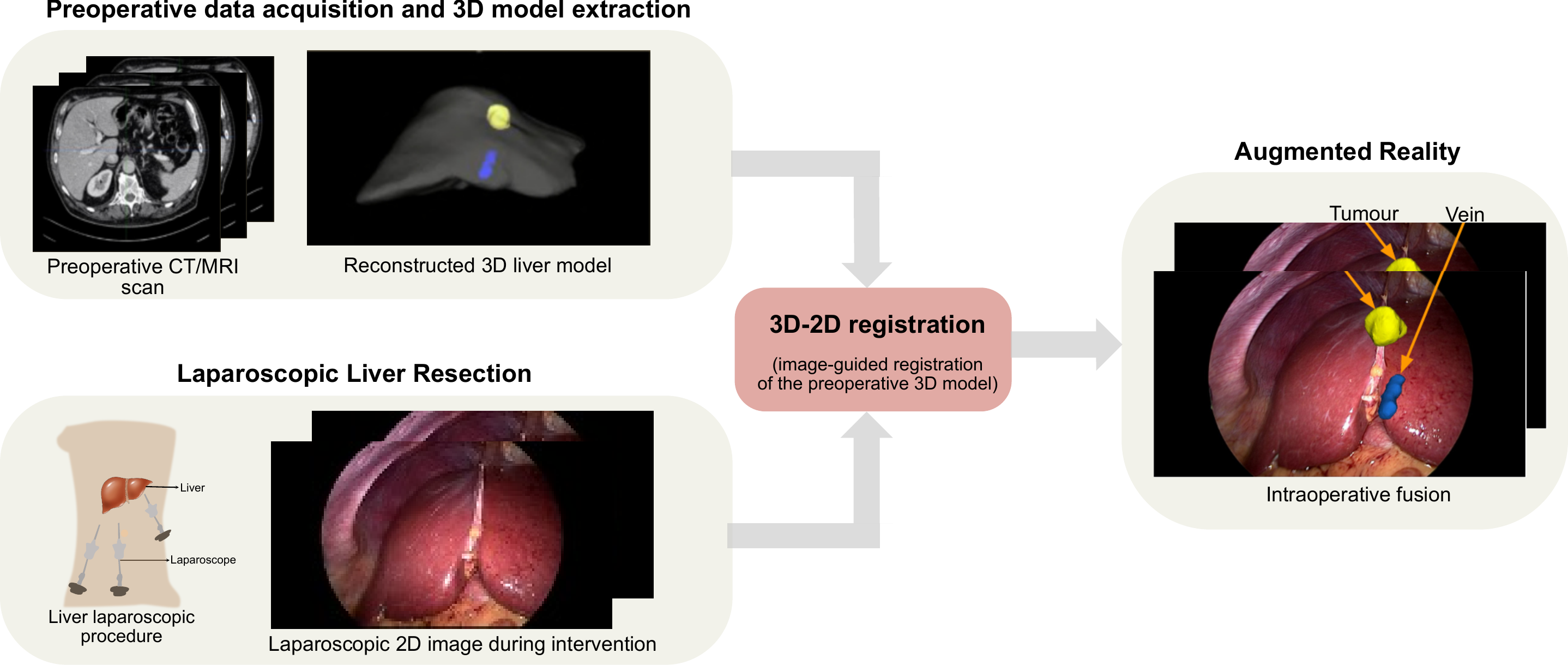}
    \caption{\textbf{Laparoscopic image fusion with preoperative 3D CT or MRI scans.} {A} preoperative 3D scan {is} first used to reconstruct {the} liver boundaries, {tumours} and major {vessels} critical for {a} safe surgery. During the {laparoscopic} procedure we overlay the {reconstructed model} using image registration, in this case 3D {meshes,} to the 2D liver view. The idea is to project 3D mesh points onto the liver boundaries that can enable understanding of the spatial location of the {tumours} and {vessels} along with the matched liver boundaries in the acquired 3D model. Such an augmented reality technique helps surgeons to locate the tumour and important landmarks during surgery. {The above results} were obtained with the semi-automatic method {from} \cite{koo2017}.}
    \label{fig:ar_pipeline}
\end{figure*}

Laparoscopic liver resection (LLR) is a minimally invasive procedure used in the removal of benign or malignant tumours. It has became increasingly popular in the last two decades {owing} to the reduced trauma to the patient and the shorter hospitalisation times. However, it remains a challenging technique due to the reduced intra-abdominal space and the lack of tactile feedback. This makes it difficult to find {intraparenchymal} structures like tumours and vessels, which increases the risk of wrong resections. Augmented Reality (AR) {could mitigate} this issue by overlaying a CT or MR-{reconstructed} 3D model onto the laparoscopic images, as shown in Figure~\ref{fig:ar_pipeline}. As depicted, the surgeons can {then} see the inner structures, and {perform} tumour resection accordingly. Owing to the liver's {substantial} flexibility, a deformable registration should be done to effectively fit the preoperative 3D model with the intraoperative data. {Once the registration is computed, the fusion can be realised.}

\begin{figure}[t!]
    \centering
    \includegraphics[width=0.50\textwidth]{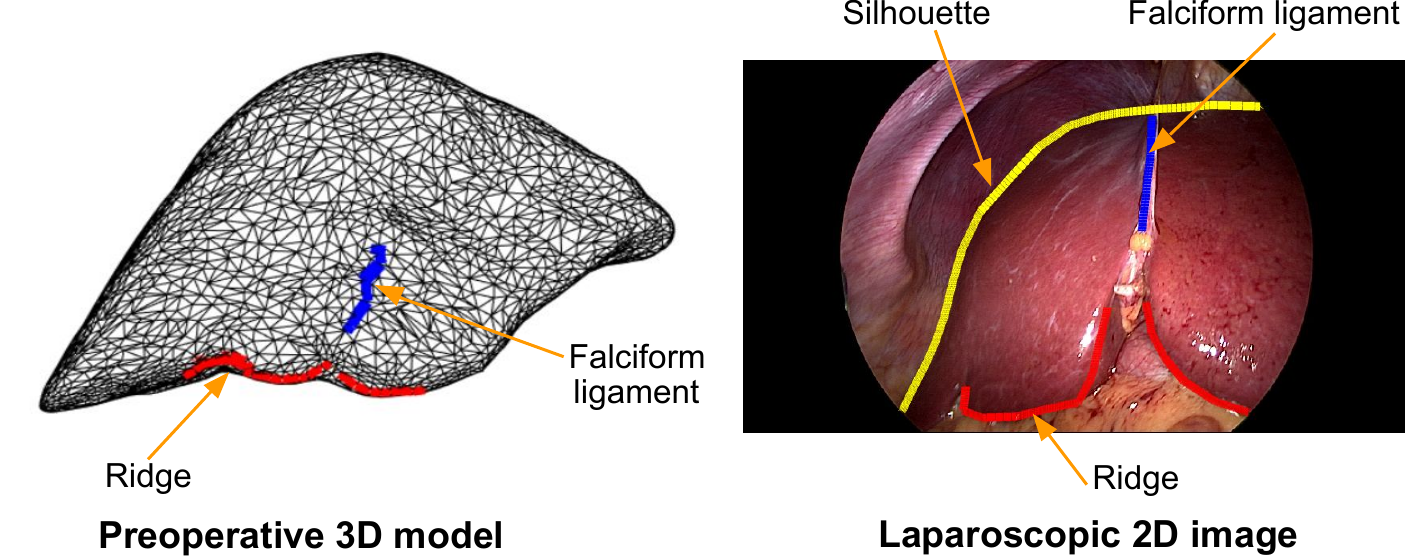}
    \caption{\textbf{Depiction of {the} 2D and 3D anatomical landmarks.} Anatomical liver landmark ground-truth annotations in the {preoperative 3D model (left)}, and in the {laparoscopic 2D image (right)}.}
    \label{fig:landmarks_liver}
\end{figure}

Existing methods register the 3D preoperative data either into 3D or 2D intraoperative data. {Most} of these methods make use of liver anatomical landmarks to constrain registration and help the preoperative model to fit in the intraoperative data. For the 3D-3D registration case, some examples {are} found in \citep{robu2018,modrzejewski2019}, where the landmarks are marked manually on both the preoperative and intraoperative 3D shapes. For the 3D-2D registration case, some examples are {found} in \citep{adagolodjo2017,koo2017,espinel2022,koo2022,labrunie2022}, where the landmarks are marked in the intraoperative images either manually or automatically, but always marked manually on the preoperative 3D models. The main problem of the 3D-3D registration methods is that they reconstruct the intraoperative data from stereoscopic cameras, which are not {always} available in surgery rooms. They may also use 3D reconstruction algorithms like Structure-from-Motion (SfM) or Simultaneous Localisation and Mapping (SLAM), which only work in rigid scenes and generally fail {for the} non-rigid liver. Therefore, we focus on the 3D-2D registration problem, which aligns the preoperative 3D models to one or several intraoperative {2D} images.\\
\indent{According} to \citep{koo2017,espinel2022}, some of the landmarks that can be used in 3D-2D registration are the liver's lifted bottom called {``ridge''}, the {``falciform ligament''}, and the top liver boundary referred to as the {``silhouette''}, {as shown in} Figure \ref{fig:landmarks_liver}. In order to accurately fit the 3D model to the images, a good correspondence between the landmarks in the laparoscopic image and the preoperative 3D model should be found. However, as the marking is usually done by hand, it will greatly depend on the user's understanding of the scene, which can be a source of inaccuracies. Moreover, the time required to manually mark these landmarks, usually several minutes, makes it difficult to integrate AR within the surgical workflow. Some of the existing methods detect the landmarks on the images automatically like the work in \cite{labrunie2022}, where a UNet network is trained and used to detect the 2D liver landmarks, while the 3D landmarks are still marked manually. The work in \cite{koo2022} also detects the 2D landmarks using a CASENet network, with the 3D landmarks being marked manually. Due to the aforementioned limitations of manual marking in terms of accuracy and time, there is a need for automating the detection of these landmarks in the image and the preoperative 3D models, as well as accurately finding the correspondences between them for the 3D-2D registration.

The Preoperative-to-Intraoperative Laparoscopic Fusion (P2ILF) challenge addresses the problem of finding the liver's anatomical landmarks in both the laparoscopic images and the preoperative 3D model, and {of} using them for 3D-2D registration. This challenge was deployed on the Grand Challenge platform \citep{p2ilf}, where the teams could register, download the training data, upload their algorithms, and run the algorithms on the test data. The challenge was divided in two phases. In phase I of the challenge, the participants {had} to detect the visible ridge, ligament, and silhouette landmarks in the laparoscopic images, and then detecting the corresponding ridge and ligament landmarks in the preoperative 3D model. In phase II of the challenge, the participants {had} to perform 3D-2D registration. For phase II, {the} participants were suggested to use the 3D and 2D landmarks detected in phase {I}. However, this was not mandatory and they were free to perform either a rigid or a deformable registration. The two phases were evaluated independently, with the quality of the detected landmarks being measured in the first phase and the registration accuracy being measured in the second phase. 

For the P2ILF challenge, surgical data was collected and annotated for 11 patients, including their corresponding preoperative 3D models, the intraoperative laparoscopic images, and the intrinsic camera parameters. A total of 208 images were annotated with the visible anatomical landmarks, along with their 3D counterparts in the 3D models. The provided data presents two main challenges: {the} drastic change in shape and appearance of the liver between patients and the limited amount of data. A total of six teams from four countries participated in the challenge. {We} describe the algorithm developed by each team and the results obtained on the test set.
{We first} present the related work, the details on the newly curated dataset for AR in LLR, the design and setup of the challenge, the approaches proposed by the participating teams, their results and insights regarding the limitations of each approach, and finally conclude with the discussions presenting {empirical and experimental hypotheses} and future work.

\section*{Related work}
\label{sec:related_work}
Registration for AR in LLR has been an active field of research {over} the last decade, with the existing methods using either monocular endoscopes, stereo endoscopes, and external devices like optical trackers and intraoperative CT scanners. Monocular methods may perform 3D-3D registration like the method in \cite{modrzejewski2019}, where an intraoperative shape of the liver was reconstructed using SfM. This shape was then combined with a set of landmarks and a biomechanical model to perform deformable registration. The registration process follows a rigid-to-deformable energy minimisation strategy, which {runs} until the convergence threshold {is} reached. Another {method} that uses SfM is presented in \cite{cheema2019}, where an intraoperative shape also serves as a target for registration. In this case, correspondences between the preoperative and the intraoperative shapes were combined with shading cues to align and deform the intraoperative shape. Similarly, \cite{espinel2022} combined the reconstructed camera poses with the anatomical landmarks and the biomechanical parameters for registration. However,  \cite{espinel2022} suggested that applying SfM in liver scenes is a difficult task due to the constant deformations and the limited range of camera movements.

Other monocular methods performing 3D-2D registration by using one or multiple images simultaneously exist in the literature. For example, a set of silhouette landmarks were manually marked in the image and combined with a biomechanical model to drive registration in \cite{adagolodjo2017}. These constraints were solved using a Gauss-Seidel iterative optimisation approach. A single-view method was presented in \cite{koo2017}, which we use as basis and motivation of our work. In this work, {the} authors combined the ridge, ligament, and silhouette landmarks with a biomechanical model to perform registration. Prior to registration, the landmarks are manually marked in both the 2D image and the preoperative 3D model. It also uses a Gauss-Seidel iterative algorithm to solve the landmark and biomechanical constraints. A set of methods that perform 3D-2D registration on multiple laparoscopic images is presented in \cite{espinel2022}, where the anatomical landmarks from all the images are combined to deal with the partial visibility problem and improve registration accuracy. 
In this case, the landmarks should be manually marked on each image separately, which increases the total registration time. In an attempt to reduce the risk of wrong annotations and the registration time, the method in \cite{koo2022} detects the anatomical landmarks in the images automatically. To achieve this, a CASENet CNN was trained with a small dataset of 133 patient images, along with a synthetic dataset consisting of 100,000 images. However, the 3D landmarks were still marked manually in the 3D model. The proposed rigid registration starts by computing a canonical liver pose, assuming that the camera is inserted close to the belly button. Then, a set of transformations were generated by randomly rotating the model about the three axes. For each of the transformations, the closest points between the 3D and 2D landmarks are found and an optimal transformation is estimated using {Perspective-$n$-Point (P$n$P)} with RANSAC. In the end, the best transformation {is} chosen based on the minimum Hausdorff distance between the 3D and 2D landmarks. Similarly, another approach that detects the landmarks automatically was proposed in \cite{labrunie2022}, where an off-the-shelf UNet was trained with 1415 laparoscopic images from 68 patients. Again, the 3D model landmarks were still annotated manually prior to surgery. The main goal of this work was to perform an initial rigid registration to serve as basis for {subsequent} deformation stages. The registration approach used a RANSAC-based {P$n$P} strategy that iteratively recomputed the correspondences between the 2D and 3D landmarks.

Methods that use stereoscopic cameras or other external devices usually perform 3D-3D registration. {In particular}, the methods in \cite{haouchine2013,soler2014,thompson2015,bernhardt2016,robu2018,luo2019} reconstruct an intraoperative 3D model of the visible liver using stereoscopic techniques. In these cases, the intraoperative 3D model is used as a target to register the preoperative 3D models. Some of the methods perform rigid registration \cite{soler2014,thompson2015,bernhardt2016,robu2018,luo2019}, while the method in \cite{haouchine2013} is the only work that performs deformable registration. {In addition to} a stereo endoscope, the method from \cite{thompson2015} also {uses} an optical tracker to locate and merge multiple stereoscopically reconstructed patches of the intraoperative liver. A major limitation of these methods is the requirement of stereo endoscopes and external tracking devices that are not commonly available in surgery rooms. Therefore, in this work we aim to find registration methods that only use the available preoperative models and a monocular endoscopic setting in the surgery room. Such methods should automatically find the liver anatomical landmarks that can then lead to 3D-2D registration automatically. We attempt to motivate {the usage of} data-driven approaches which is still {uncommon {in this problem}, as well as} to reduce both the user interactions and the registration times. {Given the high number of existing methods and a lack of unified comparison, this challenge is the first one to provide an objective comparison between registration methods for AR in LLR, which is a requirement to continue advancing in the field.}

%
\section*{Dataset}
\label{sec:dataset}
\begin{figure*}[t!]
    \centering
    \includegraphics[width=0.9\textwidth]{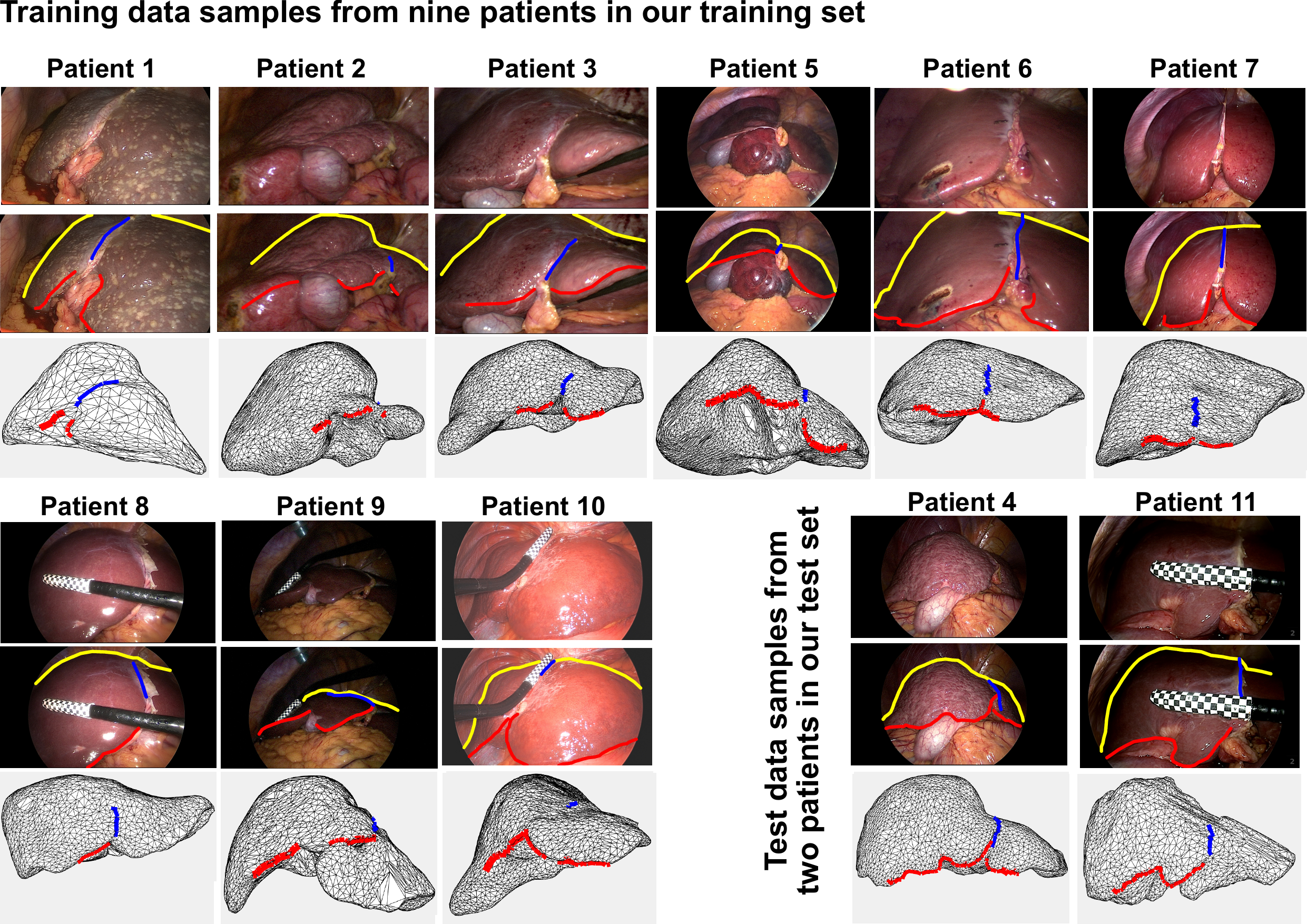}
    \caption{\textbf{P2ILF dataset:} Training and test data samples with original laparoscopic images, annotated anatomical landmarks (silhouette in yellow, ridge in red and falciform ligament in blue), and the corresponding 3D anatomical annotations (rigde in red and falciform ligament in blue) in manually aligned 3D liver {models} are provided. {The dataset contains a total of 11 patients, divided in 9 patients for training and 2 patients for testing.}}
    \label{fig:datasamples}
\end{figure*}
%
{The} training dataset is composed of {9} patients with 167 laparoscopic images, their corresponding 2D and 3D anatomical landmarks, and their respective intrinsic camera parameters. {The} test dataset is composed of {2} patients and includes 16 selected images (8 images per patient) with their corresponding intrinsic camera parameters. The training dataset was provided to the participants, who were allowed to freely split the data for training and validation. 
However, for the test phase, an online platform was used in a way that prohibited {the} teams to access the test samples directly. For the algorithmic testing reported in this paper, each algorithm was evaluated through the deployment of docker containers.

Figure~\ref{fig:datasamples} illustrates the training and test samples for the 11 patients (with one sample per patient), including the original laparoscopic images at the top, the ground-truth anatomical landmarks in the middle (with the silhouette, ridge and falciform ligament in yellow, red, and blue, respectively), and the preoperative 3D models with their corresponding 3D landmarks for the ridge (in red) and the falciform ligament (in blue). It can be observed that the appearance of the liver varies greatly across patients. Moreover, some of the patients have a visible ultrasound probe, which is common in laparoscopy as it {may be} used to identify key vessels and tumour locations during surgery. 


%
\section*{Crowd-sourced event}
\label{sec:crowdsourced_event}
%
%

\subsubsection*{Ethical and privacy aspects of the data}
\label{sec:ethical_aspects}
The preoperative and intraoperative data of this dataset were collected from {the} University Hospital of Clermont-Ferrand, France. The data collection was supported by an ethical approval with ID IRB00008526-2019-CE58 issued by CPP Sud-Est VI in Clermont-Ferrand, France. Patient consent to record data was {obtained} before each intervention. The intraoperative video streams were captured using laparoscopic cameras. 
All the collected data {were} fully anonymised before publication. In other words, no meta information (name, birth date, gender, etc.)~was passed to the participating teams. During the challenge, all participants were required to sign a data privacy statement. Redistribution or transfer of the data was strictly prohibited.~The data upon public release will be free to use (under licence CC-by-NC-SA 4.0) after the publication.
\subsubsection*{Video collection and dataset construction}
\label{sec:dataset_construction}
The dataset for the P2ILF challenge consists of two types of data: preoperative 3D liver models and {intraoperative} 2D laparoscopic images. The data were collected by the following procedure:
\begin{itemize}
    \item Several days before the liver surgery, 3D CT/MRI images of each patient were obtained.
    \item During each surgery, an exploration of the intra-abdominal scene was done in such a way that the liver was visible to the camera. A video {was} captured during the exploration. {To estimate the intrinsic camera parameters, a video of a moving checkerboard pattern was also captured with the laparoscope.}
    \item {After each surgery, 30 to 40 images were selected from the checkerboard video. Then, the images were imported into the Metashape software \protect\citep{metashape}} {and the intrinsic camera parameters were estimated. These parameters included the camera's focal length, the principal point, and the lens distortion.}
\end{itemize}

\noindent The 3D CT/MRI images were segmented using MITK, an open-source software for processing medical images. Then, the preoperative 3D models were reconstructed by interpolating the segmented images.
From the raw laparoscopic videos, the laparoscopic images were selected based on two criteria:
\begin{itemize}
    \item {Clear sharpness in terms of focusing and blur}
    \item {Noticeable} {viewpoint} change between the images
\end{itemize}

\subsubsection*{Annotation strategies and quality assurance}
\label{sec:annotation_strategy}
{Due to a lack of available LLR datasets with annotated 2D/3D anatomical landmarks, there was a need to annotate the landmarks in multiple images and preoperative 3D models. To achieve this, three of the challenge organizers, guided by the indications given by{Prof.~Emmanuel Buc, Dr.~Shahid Farid, and Dr.~Yamid Espinel, proceeded to annotate the 2D landmarks in the 183 laparoscopic images and the corresponding 3D landmarks in the 11 preoperative 3D models. From the team of annotators, Dr.~Sharib Ali and Dr.~Yueming Jin have worked extensively in artificial intelligence for surgical imaging data. Dr.~Yamid Espinel has previous experience working in AR for LLR. Once the annotations were done, they were reviewed by Prof.~Buc and Dr.~Espinel to ensure {their} quality {level and corrected if required}.} 

The annotators were required to annotate the ridge, silhouette, falciform ligament, and liver surface in every image. The tolerance error of annotation was 5 pixels. If the distance between the annotation and the actual landmark exceeded this tolerance range, it was rejected in a later review. All the annotations were done via LabelBox, an open-source collaborative web-based tool. For the annotations to be as precise as possible, the annotators were advised to use annotation tablets to perform their task. {Some important protocols that were agreed and communicated for the annotation process were}: 
\begin{itemize}
    \item {The} ligament is on the liver surface and should divide the right and left lobes
    \item {The} ridge is the curvy area located at the bottom of the liver's posterior part
    \item {The} silhouette is the occluding boundary of the liver, usually located at the upper part of the liver
    \item {The} silhouette should not go inward the ligament margin, but can go over the ligament
    \item {The} landmarks occluded by blood{, neighbouring organs,} or surgical tools should not be considered
\end{itemize}


\subsection*{Challenge tasks and setup}
\label{sec:task_setup}
We evaluated the teams on the following two tasks - a) \textbf{Task 1:} We {requested} the teams to perform 2D {and 3D} landmark segmentation on {the} laparoscopic images {and the preoperative 3D models, respectively. Landmark segmentation in the 2D laparoscopic images and in the 3D preoperative models may be seen as forming two separate tasks. However, we propose to form a single task for both types of segmentation. The rationale is that the 2D and 3D landmarks should eventually provide correspondences for later registration. Hence, it is important to allow both modalities to be segmented jointly in order for the segmentation to embody the notion of corresponding landmarks. For the 2D case, the teams were asked to segment} the ridge, the silhouette and the falciform ligament landmarks. {For the 3D case, they were asked to segment the ridge and the falciform ligament landmarks, according to the previously segmented 2D landmarks. We provided the teams with the 167 laparoscopic images, the 9 preoperative 3D models, and the corresponding 3D-2D landmark annotations from the 9 training patients. We kept the 2 test patients {undisclosed} and used their 3D-2D landmark annotations as groundtruth to assess the predictions done by the teams.} b) \textbf{Task 2:} {We {requested} the teams to register the preoperative 3D models into the intraoperative laparoscopic images, preferably by using the previously predicted 2D and 3D landmarks. This 3D-2D registration could be either rigid or deformable. We used the 2D ridge, the falciform ligament, and the silhouette landmarks from the 2 test patients as groundtruth to assess the registrations done by the teams.} 

The input and output data to be used in each of the tasks are shown in Figure~\ref{fig:docker_algorithmic_evaluation}. The teams were required to run their methods in a docker-based deployment framework{, hosted in the Grand Challenge platform~\protect\citep{grandchallenge}}. The test data was not accessible by the teams. Another docker-based container was developed to assess the predictions and generate the evaluation metrics automatically. 

\begin{figure}[t!]
    \centering
    \includegraphics[width=0.49\textwidth]{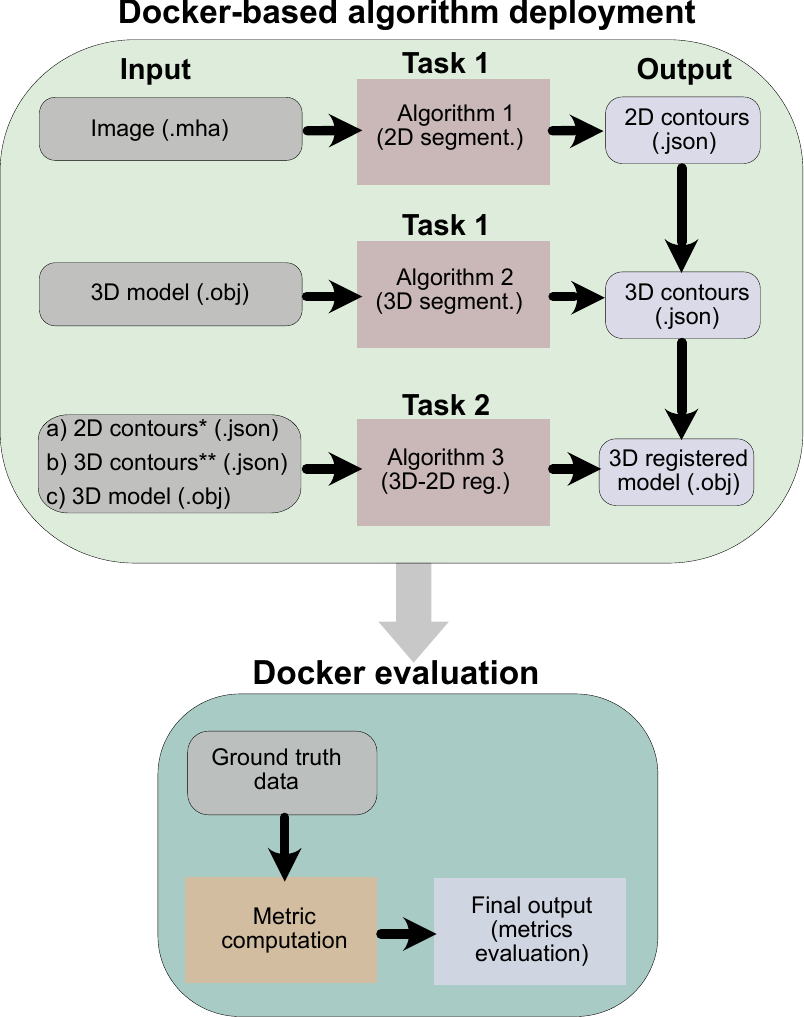}
    \caption{\textbf{Submission procedure of the P2ILF Teamchallenge:} A docker container system for submission was established on {the Grand Challenge} platform. Each liver model and corresponding images together with intrinsic camera parameters were provided to the challenge participants. The algorithmic submission required different inputs for {the} prediction of 2D liver landmarks, 3D liver landmarks, and the use of these landmarks for registration of {the} 3D model to the laparoscopic images. Finally, the outputs from each team's algorithm were evaluated using different metrics {(see the section Evaluation Metrics for more details)}.}
    \label{fig:docker_algorithmic_evaluation}
\end{figure}
%
%
\section*{Team methods}
\label{sec:team_methods}
{We describe} the methods proposed by each participating team. We describe how every method deals with each of the two tasks, namely the landmark detection task and the 3D-2D registration task. At the end of the section, we provide {in table \protect\ref{table:methods_summary_task1}} a summary of the 2D-3D landmark segmentation strategies proposed for Task 1 and in table \ref{table:methods_summary_task2} the 3D-2D registration strategies proposed for Task 2. {The participating teams in the challenge {were} the \textit{BHL} team from Fudan University (China), the \textit{UCL} team from University College London (United Kingdom), the {\textit{GRASP}} team from the University of Pennsylvania (United States), the \textit{VOR} team from the University of Science and Technology of China (China), the \textit{NCT} team from the National Center for Tumour Diseases in Dresden (Germany), and the \textit{VIP} team from the Hong Kong University of Science and Technology (China).}

\subsection*{\textit{Team 1} (BHL team)}
\label{sec:team_bhl}
The BHL team has proposed an automatic way of detecting the 2D and 3D landmarks using deep learning methods for the first task and a classical semi-automatic rigid registration approach that uses the detected landmarks for the second task.

\subsubsection*{Task 1}
For the segmentation of the 2D anatomical landmarks, photometric and geometric distortions were used to augment the training dataset, including variations in brightness, contrast, random noise, scaling, cropping, clipping, and rotation. The labels were extended by three pixels using an adjacent pixel strategy, and the images were resized to $256\times512$ pixels for GPU acceleration purposes. After applying Fast Fourier Transform on the original images, Inverse Fourier Transform was applied on the high-frequency components to obtain contour-enhanced images \citep{yang2020fda}. The team found that this contour enhancement improved segmentation of the silhouette and the ligament landmarks, but not of the ridge landmark. Therefore, two separate ResUnet \citep{zhang2018road} models were used. One {ResUnet} segmented the ridge landmark from the original image, and the other {one} segmented the silhouette and the ligament landmarks from the contour-enhanced image. The labels were then {dilated} by three pixels. The output of the models {were} constrained with a Dice loss and a CE loss, with {an} L1 loss $L_B$ being introduced to constrain the consistency of the two models:

\begin{equation}\label{eq:lossL1}
    L_B = |m_1 - m_2|,
\end{equation}
\noindent where $\{m_1,m_2\}$ are the output maps of the first and second Res-UNet, respectively.

To segment the anatomical 3D landmarks, a coarse-to-fine point cloud segmentation strategy {was} designed to compensate the impact of data augmentation and class imbalance, mainly due to the low number of points that the landmarks have compared to the rest of the mesh. {The} dataset was augmented by applying random rotations and scales ($0.75$ to $1.25$ times) on the 3D models to mimic the liver's size and orientation. To {deal with the} class imbalance {problem} in the mesh data, {the groundtruth landmarks were dilated twice using a distance threshold of 20 mm.} The vertices in each mesh were then normalised as follows:
\begin{equation}\label{eq:vertices}
    (x,y,z) = \left(\frac{x_i-x_{mean}}{x_{max}-x_{min}},\frac{y_i-y_{mean}}{y_{max}-y_{min}},\frac{z_i-z_{mean}}{z_{max}-z_{min}}\right),
\end{equation}
\noindent{where} $mean$ is the average of coordinates of all vertices, $max$ and $min$ are the maximum and minimum coordinates,  respectively.~After normalisation, the point cloud dataset was used to train a PointNet++ network \citep{qi2017pointnet++} {to} segment the ridge and the falciform ligament.
\subsubsection*{Task 2}
{The task of} 3D-2D registration was achieved by first taking the end points of the predicted 2D and 3D ridge landmarks, and then by manually sampling the 2D and 3D landmarks to have the same number of points. These point correspondences served as input to the P$n$P solver from {the} OpenCV library, along with the intrinsic camera parameters. The obtained rigid transformation, described by a rotation $R \in SO(3)$ and a translation $t \in \mathbb{R}^3$, is used to register the 3D model into the image. 

\begin{figure*}[!t]
    \centering
    \includegraphics[width=1.0\textwidth]{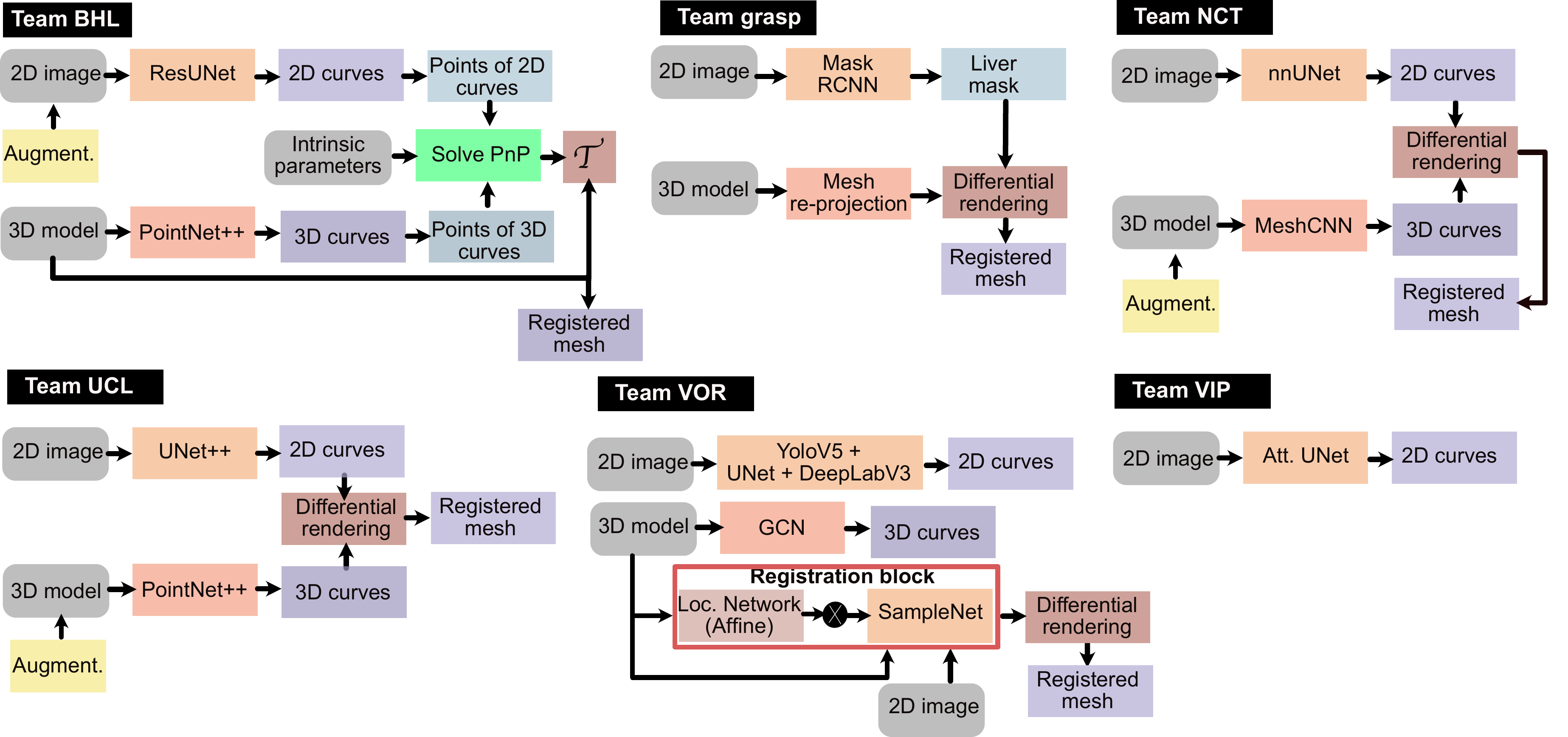}
    \caption{\textbf{General pipeline of the six team methods.} \textbf{Team BHL:} The input 2D image and 3D model are first processed and augmented. Two ResUNets are used to segment the 2D landmarks in the images, and one PointNet++ is used to segment the 3D landmarks in the {preoperative 3D model}. {To perform 3D-2D registration, the} correspondences are fed to the P$n$P algorithm and a transformation matrix is obtained. \textbf{Team {GRASP}:}  Mask-RCNN is used to {generate a} 2D mask of the liver, which is then used to {perform 3D-2D registration} by minimising a silhouette reprojection error {through differentiable rendering}. \textbf{Team NCT:} nnUNet and MeshCNN are used {to segment the 2D and 3D landmarks, respectively.} Differential rendering is {then} used to {perform 3D-2D registration by minimising a reprojection error of the previously detected landmarks}.
  \textbf{Team UCL:} UNet++ is used {to segment the} 2D landmarks, while PointNet++ is used {to segment the} 3D landmarks. This team also used differential rendering {to perform 3D-2D registration}.
   \textbf{Team VOR:} {The} 2D case is treated as a pixel segmentation task and the 3D case as a vertex classification task. Differentiable rendering is {then} used to {perform 3D-2D registration by} generating 2D images from the affine transformations computed by the localisation networks. The shape regularisation terms provide extra supervision to avoid undesired mesh deformations. \textbf{Team VIP:} The team only participated in task 1. Attention UNet was used for the pixel segmentation task of the anatomical liver landmarks in the laparoscopic images.}
    \label{fig:bhl_pipeline}
\end{figure*}
\subsection*{{\textit{Team 2} ({UCL} team)}}
\label{sec:team_ucl}
{The UCL team} has also proposed an automatic way of detecting the 2D and 3D landmarks using deep learning methods for the first task, with an automatic rigid registration approach based on differential rendering for the second task.

\subsubsection*{Task 1}
The proposed method uses UNet++ \citep{zhou2018unet++} to segment the 2D anatomical landmarks. A combination of Dice loss and Hausdorff distance was used as global loss for training. To complement the provided training dataset, synthetic liver images were generated using Unity and Blender. 100,000 images were generated in Unity, using 3D models {and} textures purchased from the Unity Asset Store \citep{unity_store}, and textures taken from freely available sources. 
For each iteration of {the} simulation, random values were uniformly sampled for texture, camera position, lighting effects, motion blur and lens distortion. For each {of the 9} patients in the training set, 1000 extra images were simulated in Blender using the patient-specific liver models. 
Other structures like fat and background were manually generated. All textures were procedurally generated and randomised at simulation time, with variable camera positions and lighting effects at each iteration. The UNet++ was pretrained using the synthetic data for 10 epochs with an ADAM optimiser and an adaptive learning rate starting from $10^{-6}$. Then, the model was fine-tuned using the patient data for 10 epochs.


The 3D landmarks were segmented using geometric deep learning including PyTorch Geometric and PointNet++. A global loss combining {the} Hausdorff distance and Negative Loss Likelihood (NLL) was used for training. Spectral augmentation \citep{foti2020intraoperative} {was} performed to produce a broader collection of 3D models. {In addition to the 9 preoperative 3D models provided in this challenge, the team also used the phantom model from \protect\cite{espinel2022} for training purposes. For 9 of these models,} 199 augmentations were generated, giving a total of 1800 models for training. {The remaining patient was also augmented with 199 extra models, giving a total of} 200 models for validation. Training was conducted over 1000 epochs with an ADAM optimizer, using a learning rate of $10^{-3}$.

\subsubsection*{Task 2}
For the 3D-2D registration task, the team proposes a differential rendering pipeline using PyTorch3D. This pipeline iteratively renders the silhouette of the preoperative liver model $M$ and reduces a registration loss. The position of the liver model is initialised with a rotation $R = I_3$, where $I_3$ is a $3 \times 3$ identity matrix, and a translation $t = (0, 0, 500)$. An initial registration process is carried out over 100 iterations, where every iteration is performed in five steps. First, the silhouette of the 3D liver is rendered based on the current $R$ and $t$. Second, an image loss is computed between the rendered 3D silhouette and the 2D silhouette detected in the first task. Third, all the points of the 3D ridge and ligament landmarks detected in the first task are projected in 2D. Fourth, a Chamfer loss is computed between the projected 3D landmarks and the corresponding 2D landmarks. Fifth, the image and Chamfer losses are backpropagated through the network to update $R$ and $t$.

After the first 100 iterations, a rough initial alignment is achieved, which is used to identify the point correspondences between the 3D and 2D landmarks extracted from the first task. After this, another 25 iterations are carried out, with the difference that only the 3D point correspondences found in the initial alignment are used.

\subsection*{\textit{Team 3} ({GRASP} team)}
\label{sec:team_grasp}
The {GRASP} team proposed a registration method based on automatic liver segmentation that includes two stages: first, the liver region is segmented with a Mask R-CNN network \citep{he2017mask} and second, the preoperative 3D model is rigidly registered by using the previously segmented liver mask, by minimizing a silhouette reprojection error with a classical method.

\subsubsection*{Task 1}
The team segments the liver region by using a Mask R-CNN pretrained on the COCO instance segmentation dataset \citep{lin2014microsoft}. The weights for the non-liver classes where then removed, leaving only liver and background. The model was then fine-tuned on the provided patient dataset for 15 epochs. To prepare the training dataset, they annotated the visible liver on the patient images using the \textit{Labelme} annotation tool. Images from 7 patients were used for training and from 2 patients for testing.

\subsubsection*{Task 2}
An optimisation-based approach is used to rigidly register the preoperative 3D models to the laparoscopic images. A transformation {$T$} registers the preoperative 3D model  $M$ using a rotation $R$ and a translation $t$. The optimal transformation $T^*$ is computed for every image by minimising the silhouette reprojection error:

\begin{equation}\label{eq:reprojection_error}
    T^{*} = \argmin_{T} E(T,M,S),
\end{equation}

\noindent where $E(T,M,S)= L_{\epsilon}(D(T(M))-S)$ is the reprojection error function, $\lambda$ is a weighting term, $L_{\epsilon}$ is the smooth $L_1$ loss, $D$ is the differential rendering function \citep{ravi2020accelerating}, and $S$ is the predicted liver mask. 

The optimization process begins by initializing the mesh to a canonical pose, i.e., estimating the position of the organ with respect to the camera. This canonical pose is calculated by manually registering meshes to 15 images and taking the average of all the poses. Then, the optimisable parameter ${T_j}$ is created in Pytorch. Using Pytorch3D's differentiable rendering module, for each optimization step $j$ a silhouette image is rendered using the 3D model transformed by ${T}_{j}$. By back-propagating the loss between the rendered silhouette and the predicted silhouette from Task 1, a new pose ${T}_{j+1}$ is computed. The 3D model is then registered in the next step using this new pose. Every 200 steps takes roughly 30 seconds. 


\subsection*{{\textit{Team 4} ({VOR} team)}}
\label{sec:team_vor}
{The VOR} team {detects} the 2D and 3D landmarks through a multi-staged deep learning approach, and perform rigid registration by combining deep learning with differentiable rendering.

\subsubsection*{Task 1}
The 2D landmark segmentation case is treated as a pixel segmentation task and the 3D case as a vertex classification task. Concretely, the team suggested a multi-staged network for each type of anatomical landmark that incorporates a pre-trained UNet \citep{ronneberger2015u} for initial segmentation and a YOLOv5 \citep{redmon2016you} as a region proposal module. The anatomical detection from the UNet is first converted to a box-shaped segmentation mask. This mask is then combined with the results from the YOLOv5 to form a region-of-interest (ROI) from where representative features are learned for the final segmentation. Then, this ROI mask is multiplied with the original RGB image, and the resulting patch is downsampled from $1920\times1080$ pixels to $960\times540$ pixels. A DINO transformer \citep{caron2021emerging} is used to generate feature representations from the previously generated patches. Finally, a DeepLabV3 network \citep{chen2017rethinking} is implemented to detect the final anatomical curves. For the 3D landmark segmentation, a Graph Convolutional Network (GCN) \citep{kipf2016semi} is chosen to perform mesh vertex segmentation. A random vertex masking is used for augmentation purposes, along with a mesh normalisation that converts the data into unit space to improve training stability. In order to train the segmentation networks, the dataset was split in 80\% for training and 20\% for validation purposes. The team verified that each set contained images from all {of} the patients. 


\subsubsection*{Task 2}
The proposed 3D-2D registration method uses sampling-based localisation networks. 
It is designed to deal with two main problems. First, correlating the 2D image with the 3D mesh and, second, preserving the mesh topology and volume during registration. To train the localisation networks, a 2D image similarity loss is combined with a 3D shape regularisation term. Concretely, the localisation networks are inspired by the Spatial Transformer network \citep{jaderberg2015spatial}. They learn a parameterised affine transformation $T$ at every stage, which is then applied to the preoperative liver model $M$. Then, a sampling module projects the visible vertices onto the images and associates the projected vertices with {colours}. A Soft Rasterizer \citep{liu2019soft} generates an image from the projected vertices in order to compute the image similarity loss. The affine transformations learned by the localisation networks are represented as seven-dimensional vectors with three channels for rotation, three for position, and one for scale. The 2D loss exerts the major supervision when learning the optimal transformation $T^*$ for registration. This dimensional reduction inevitably loses the control of 3D shape properties.  
To compensate the impact of such reduction, two shape-based regularization terms were added: {a} Laplacian loss that quantifies the smoothness of the local surface around each vertex, and an edge length loss that penalises significant changes in the edge lengths, avoiding undesired deformations in the mesh such as flattening, erosion, or dilation. In addition, the team observed that decoupling the transformation prevents the model to generate unsuitable affine transformations. 


\subsection*{{\textit{Team 5} (NCT team)}}
\label{sec:team_nct}
The NCT team used UNet-like networks to solve the 2D and 3D landmark segmentation steps of Task 1. They also used differentiable rendering to perform 3D-2D rigid registration in Task 2 by a classical method.

\subsubsection*{Task 1}
For the 2D landmark segmentation, the team adopted the nnUNet network \citep{isensee2021nnu}. Since the ground-truth labels were made up of very thin lines, they slightly increased the width of the labels by dilating them by up to 10 pixels. Training is performed using a five-fold cross-validation scheme, which results in five sets of network weights. Since each of the networks generate under-segmented results, they are combined as the union of all the predicted ligaments, silhouettes and ridges.

For the 3D landmark segmentation, a UNet-like \citep{ronneberger2015u} with MeshCNN network~\citep{hanocka2019meshcnn} incorporating skip connections and residual blocks was used. MeshCNN operates directly on the triangular meshes, extracting local edge features to make predictions that are invariant to rotation, translation, and scale of the input data.  An important characteristic is the task-adaptive pooling of the mesh to lower resolutions, which is achieved by collapsing edges according to the scale of the learned feature representations. Two MeshCNN networks are trained on synthetic data in a supervised manner. They infer edge-wise class predictions which are then merged to form the final landmark segmentations. {Labelled} synthetic data were generated from the released patient models in two steps. First, the ridge and ligament 3D landmarks from all views are merged for each patient. Second, the meshes {were} downsampled to improve performance and then deformed using finite element simulations, adjusting the labels accordingly. A total of 8208 annotated deformations were generated to form up the training dataset. The original undeformed patient and phantom meshes were used for validation. In order to tackle the class imbalance problem, {the} networks {were} trained using a cross entropy loss with the highest weight assigned to the falciform ligament class. {The hyperparameters were} tuned to {favour} results that {were} suitable for the downstream registration task. The team observed that MeshCNN performs better at ridge or ligament prediction at different stages of training. Early trained models predicted the ridge more reliably, while later trained models appear to place the ligament at a more reasonable position. Therefore, two sets of weights are used to predict the two classes independently. The union of the two predictions completes the final landmark segmentation. Once the edges have been segmented, each vertex is assigned the class of the edges it is a part of. This way, vertices can belong both to the ridge and ligament classes. Prior to any operation, MeshCNN normalises edge lengths by their mean and standard deviation in the dataset. For the normalisation step during inference on unseen test samples, the mean and standard deviation of the original patient data is used.

\subsubsection*{Task 2}
The 3D-2D registration task is solved through an optimisation scheme based on differentiable rendering. The system renders the liver mesh and the detected ridge and ligament landmarks using a virtual camera. The rendered 3D landmarks are then compared with the segmented 2D landmarks, and a pixel-level mean squared error is obtained. This error is back-propagated and gradients of rotation and translation are calculated. The gradients are then used to update the rotation $R$ and translation $t$ of the preoperative liver model $M$. Finally, a new render of the registered model $M$ is made. This process is repeated several times, resulting in an iterative 3D pose-optimization scheme using only 2D pixel-level losses. To compute the mean squared error loss, pixels belonging to the ligament are weighted by a factor of $5$, to the ridge by a factor of $1$, and to the silhouette by a factor of $0.5$. This is done because the ligament has the strongest influence on the liver's orientation while covering the least amount of area, and the silhouette is view-dependent and may give incorrect information during early iterations.

The rendering process can be divided into four steps. First, colours are assigned to the mesh vertices based on the 3D landmark segmentation results, for which the ridge vertices are {coloured} in red and the ligament vertices in blue. Second, Pytorch3D \citep{ravi2020accelerating} is used to render the mesh and the 3D landmarks using the intrinsic camera parameters, resulting in an image that shows only the ridge and the ligament landmarks. Third, the silhouette landmark is added by rendering the liver mesh again using a single fixed color. Then, an edge-detection filter on the vertical direction extracts the boundary gradient of the liver. The pixels for which the gradient is negative (transition from the background to the liver) are considered as part of the silhouette and a contour is extracted. Fourth, the silhouette landmark is dilated to create a thicker curve, and then combined with the ridge and ligament landmarks.

Before starting the registration optimisation, the initial pose of the preoperative liver model is set at a random location on the camera's optical axis, i.e. in the positive $z$ axis of the world reference frame. An initial rotation is set so that the liver’s anterior side faces the camera, and then a random rotation of less than $90^{\circ}$ is applied on every axis. The camera is kept fixed at the origin throughout the whole procedure and only the liver is translated and rotated. The liver scale is also kept fixed. After the first render is done and to speed up convergence, the position of the rendered 3D ligament is compared to the position of the segmented 2D ligament. The 3D liver is then translated parallel to the camera plane until the two ligaments overlap in the rendered image.

The optimization process is run for 150 iterations using stochastic gradient descent at a linearly decaying learning rate. The rest of the scene parameters are not modified during the optimization. This process is repeated for 30 different random pose initialisations to increase robustness against bad initial alignments. To speed up the process, the laparoscopic and the rendered images were scaled to one fifth of their original size. 


%
\subsection*{{\textit{Team 6} (VIP team)}}
\label{sec:team_vip}
The VIP team only participated in the 2D landmark segmentation step of Task 1, with a method that detects the landmarks by combining a UNet network with attention gates.

\subsubsection*{Task 1}
The team proposed a 2D landmark segmentation method that uses Attention UNet~\citep{oktay2018}. This structure integrates Attention Gates (AG) to the UNet to reduce false-positive predictions in irrelevant structures. AGs not only strengthens features of salient boundaries passing through the skip connections but also filter out gradients from unrelated regions during both forward and backward passes. 
Before training the network, the released training dataset was augmented by applying random flip, Gaussian noise, Gaussian blur, and light adjustment to mitigate the over-fitting issue. The dataset was randomly split into training and validation sets with a ratio of 4:1. A cross-validation strategy was followed to select the best checkpoint for inference.


The framework was implemented in PyTorch and trained using an NVIDIA RTX 2080 GPU. Before training, the images were resized to 272x480 pixels and then fed into the Attention UNet. Images in each mini-batch were randomly sampled from different patients to ensure diversity. An Adam optimizer \citep{kingma2014} was used for training, along with a cross-entropy loss combined with an IoU loss. Following \cite{oktay2018}, the gating parameters were initialised so that the attention gates pass through feature vectors at all spatial locations. The network was trained from scratch for 50 epochs with an initial learning rate of $10^{-4}$ and a batch size of 16. The learning rate was then decreased by 0.9 after every 5 epochs.

\begin{table*}[!h]
\caption{Summary of the participating teams in \textbf{Task 1} of the P2ILF Challenge (2D-3D landmark segmentation)\label{table:methods_summary_task1}}
\centering
\small
\begin{tabular}{l|l|l|l|l|l|l|l|l|l|l} 
\hline
\multirow{2}{*}{\textbf{Team}} & \multicolumn{2}{c|}{\textbf{Algorithm}}                                                  & \multicolumn{2}{c|}{\textbf{Loss function}}                                                   & \multicolumn{2}{c|}{\textbf{Preprocessing~}}                                                                                                        & \multicolumn{2}{c|}{\textbf{Data aug.}}                                                 & \multicolumn{2}{c}{\textbf{Pretraining}}            \\ 
\cline{2-11}
                                    & \textbf{2D}                                                    & \textbf{3D}                                                    & \textbf{2D}                                           & \textbf{3D} & \textbf{2D}                                            & \textbf{3D}                                                      & \textbf{2D}                                             & \textbf{3D} & \textbf{2D} & \textbf{3D}  \\ 
\hline
BHL                                 & ResUnet                                                                     & PointNet++                                                                  & \begin{tabular}[l]{@{}l@{}}DSC+CE\\+$l_1$  \end{tabular}                                        & CE  & \begin{tabular}[l]{@{}l@{}}FFT, IFT, \\ dilation~\end{tabular} & \begin{tabular}[l]{@{}l@{}}Mesh \\norm.\end{tabular}                  & \begin{tabular}[l]{@{}l@{}}Yes\end{tabular} & R/S                 & No                     & No                      \\ 
\hline

{GRASP}                         & \begin{tabular}[l]{@{}l@{}}Mask \\ RCNN  \end{tabular}                                                             & No                                                                        & \begin{tabular}[l]{@{}l@{}}Log+$l_1$\\+ CE  \end{tabular}           & No                     & \begin{tabular}[l]{@{}l@{}}Liver mask \\annotation\end{tabular}     & No                                                                          & No                                                                 & No                     & COCO                     & No                      \\ 
\hline
NCT                                 & nnUNet                                                                     & MeshCNN                                                                     & CE+DS                                                         & CE                    & Dilation                                                      & \begin{tabular}[l]{@{}l@{}}Label merge,\\Mesh norm.\end{tabular} & Yes                                                                 & ST                     & No                     & No                      \\ 
\hline

  UCL                 & UNet++                                                                      & PointNet++                                                                  & \begin{tabular}[c]{@{}c@{}}DSC\\+Hfd \end{tabular} & \begin{tabular}[l]{@{}l@{}}  Hfd \\ +NLL \end{tabular}           & \begin{tabular}[l]{@{}l@{}}  Synthetic data \\ generation \end{tabular}                                                                 & No                                                                          & \begin{tabular}[l]{@{}l@{}}Yes\end{tabular}     & SP                     & ST                & No                      \\ 
\hline
VIP                                 & Att. UNet                                                             & No                                                                        & CE + IoU\footnotemark[\value{footnote}]                                                           & No                     & Resizing                                                            & No              &Yes                                                       & No                     & No                     & No                \\
\hline
VOR  & \begin{tabular}[l]{@{}l@{}}Various\end{tabular} & \begin{tabular}[l]{@{}l@{}}GCN\end{tabular} & CE   & CE             & No       & \begin{tabular}[l]{@{}l@{}}Mesh norm.\end{tabular} & No   & VM   & No   & No  \\
\hline
\multicolumn{11}{l}{\footnotesize{Various: UNet + YOLOv5 + DINO +~DeepLabV3; Att.: Attention; CE: Cross-Entropy; FFT: Fast Fourier Transform;}} \\ 
\multicolumn{11}{l}{\footnotesize{IFT: Inverse Fourier Transform; Hfd: Hausdorff distance; DSC: Dice similarity coefficient; IoU: Intersection over union; norm.: normalisation;}} \\
\multicolumn{11}{l}{\footnotesize{S: scaling, R: rotation; ST: synthetic data; ST: synthetic data; SP: Spectral augmentation; VM: Vertex masking}}
\end{tabular}
\end{table*}
%
\begin{table*}[th!]
\centering
{
\caption{Summary of the participating teams in \textbf{Task 2} of the P2ILF Challenge (3D-2D registration)}
\begin{tabular}{l|l|l|l|l} 
\hline
\textbf{Team} & \textbf{Algorithm}                                                                                    & \textbf{Registration constraints}                                               & \textbf{Loss function}                                                                       & \textbf{\textbf{Type}}  \\ 
\hline
\hline
BHL                & Iterative {P$n$P}                                                                                         & 3D-2D Ridge                                                                     & 2D reprojection error                                                                        & Rigid                                   \\ 
\hline
{GRASP}        & \begin{tabular}[l]{@{}l@{}}Iterative Differentiable Rendering\end{tabular}                           & 2D Silhouette                                                                   & 2D reprojection error                                                                        & Rigid                                   \\ 
\hline
NCT                & \begin{tabular}[l]{@{}l@{}}Iterative Differentiable Rendering\end{tabular}                           & 3D-2D Ridge + Ligament                                                          & 2D reprojection error                                                                        & Rigid                                   \\
\hline
UCL        & \begin{tabular}[l]{@{}l@{}}Iterative Differentiable Rendering\end{tabular}                          & \begin{tabular}[l]{@{}l@{}}3D-2D Ridge + Ligament\\+ 2D Silhouette\end{tabular} & \begin{tabular}[l]{@{}c@{}}2D image similarity\\+ 2D Chamfer loss\end{tabular}               & Rigid                                   \\ 
\hline
VOR         & \begin{tabular}[l]{@{}c@{}}Multi-staged Spatial Transformers\\+ Differentiable Rendering\end{tabular} & Visible liver surface                                                           & \begin{tabular}[l]{@{}l@{}}2D image simialrity\\+ 3D shape-based \\ regularisation\end{tabular} & Rigid                                   \\ 
\hline
\hline
\end{tabular}
\label{table:methods_summary_task2}
}
\end{table*}
%
\section*{Results}
\label{sec:results}

\subsection*{Evaluation metrics}
\label{sec:metrics}

The metrics used to evaluate the tasks vary according to the nature of the problem to be solved. {Task 1 uses} Precision {Dice Coefficient,} and Symmetric Distance \citep{francois2020} to {assess} the predicted 2D landmarks{, along with} 3D Chamfer Distance to {assess} the predicted 3D landmarks. Task 2 {uses the} 2D Hausdorff Distance to measure the accuracy of 3D-2D registration.

\subsection*{Task 1: Metrics for assessing the 2D and 3D landmark segmentation tasks}

%
\subsubsection*{Precision}
We use precision $P$ to measure the quality of the predicted 2D landmarks at a pixel level. It corresponds to the number of true positives over the total number of predicted pixels (true positives and false positives). {Precision is a commonly used metric in semantic segmentation to evaluate the quality of the predictions \protect\citep{taha2014}:}
\begin{equation}\label{eq:metric_precision}
    P = \frac{\abs{TP}}{\abs{TP}+\abs{FP}},
\end{equation}

\noindent where $TP$ are the true positives and $FP$ are the false positives.

%





%

\subsubsection*{Dice Coefficient}
We use Dice coefficient $DSC$ to measure the similarity between the predicted and the ground-truth landmarks. It corresponds to the intersection of the pixels in the predicted and ground-truth landmarks, over the total number of pixels in both landmarks. {Dice coefficient is also a commonly used metric in semantic segmentation to evaluate the accuracy of the predictions \protect\citep{muller2022}:}
\begin{equation}\label{eq:metric_precision}
    DSC = \frac{2 \abs{B_{I} \cap C_{I}}}{\abs{B_{I}} + \abs{C_{I}}},
\end{equation}
\noindent where $B_I$ is the set of predicted image landmarks and $C_I$ is the set of ground-truth image landmarks.

\subsubsection*{Symmetric Distance}
We use the Symmetric Distance score proposed in \cite{francois2020} to assess the similarity of the predicted and ground-truth landmarks. This score takes five performance criteria into account. First, the ground-truth landmarks should not be missed and there should be no spurious predictions. Second, the predictions should be close to the ground-truth landmarks. Third, each ground-truth landmark should only produce a single prediction. Fourth, the score should be invariant to the image resolution. Fifth, the score should be invariant to the amount of ground-truth landmarks. The score is thus defined as:

\begin{align}\label{eq:metric_symmetric_distance}
    G = \frac{1}{2|C_I|d_{max}} \left( \sum_{b_I \cap Q} d_S(b_I,c_I \setminus FN) + \sum_{c_I \setminus FN} d_S(c_I,b_I \cap Q) \right)  \nonumber \\ \!\!\!\!\!\!\!\!\!\!\!\!\!\!\!\!\!\!\!\!\!\!\!\!\!\!\!\!\!\!\!\!\!\!\!\!+\frac{|FP|}{|I|-2|C_I|d_{max}} + \frac{|FN|}{|C_I|},
\end{align}

\noindent where $G$ is the symmetric distance score, $d_{max}$ is a tolerance distance that defines if a predicted landmark is spurious or not, $b_I \in B_I$ is a landmark in the set of predicted image landmarks $B_I$, $c_I \in C_I$ is a landmark in the set of ground-truth image landmarks $C_I$, $Q$ is the tolerance region around the ground-truth image landmarks defined by $d_{max}$, $FP$ and $FN$ are the false positive and the false negative predictions, respectively, and $d_S()$ is a symmetric distance function.

%

%

%



%
\subsubsection*{3D Chamfer Distance}
We measure the similarity between the predicted and ground-truth 3D landmarks by means of a 3D Chamfer Distance. It corresponds to the sum of the squared distances between the nearest neighbour correspondences of the predicted and ground-truth landmarks. {The 3D Chamfer Distance {$d_C$} is a standard metric used to measure the similarity and completion between two point clouds \protect\citep{wu2021}:
\begin{equation}\label{eq:metric_3D_chamfer}
    d_C(v,w) = \frac{\sum_{v} \min\limits_{w} \norm{v-w}^2}{|v|} + \frac{\sum_{w} \min\limits_{v} \norm{v-w}^2}{|w|},
\end{equation}

\noindent where $v \in b_M$ are the points in a predicted model landmark $b_M$, and $w \in c_M$ are the points in the corresponding ground-truth model landmark $c_M$.

We use the average Chamfer distance $F$ between the predicted and ground-truth landmarks as the evaluation metric:
\begin{equation}\label{eq:metric_nearest_neighbor}
    F = \frac{1}{|B_M|} \sum_{b_M} d_C(b_M,c_M),
\end{equation}

\noindent where $B_M$ is the set of predicted model landmarks.

%




\begin{landscape}
\begin{table}[t!]
\centering
\caption{\textbf{Evaluation of 2D segmentation of landmarks using region-based metrics:} Precision and dice coefficient scores (DSC) are provided for 16 images from the 2 test cases (patients 4 and 11). Each evaluation metric includes values for the ridge, the falciform ligament, and the silhouette landmarks. The higher the precision and DSC values the better. The best results are in bold, the second best are underlined.}
\begin{adjustbox}{width=1.10\columnwidth,center}
\begin{tabular}{c|cc|cc|cc|cc|cc}
\hline
\multirow{1}{*}{\textbf{Test}} & \multicolumn{2}{c|}{\textbf{BHL}}                                                        & \multicolumn{2}{c|}{\textbf{NCT}}                                                        & \multicolumn{2}{c|}{\textbf{UCL}}                                                        & \multicolumn{2}{c|}{\textbf{VIP}}                                                        & \multicolumn{2}{c}{\textbf{VOR}}                                                        \\ \cline{2-11} 
                               {\textbf{samples}}    &  \textbf{Precision} & \textbf{DSC} & \textbf{Precision} & \textbf{DSC} & \textbf{Precision} & \textbf{DSC} & \textbf{Precision} & \textbf{DSC} & \textbf{Precision} & \textbf{DSC} \\ \hline
4\_3              & 0.14/0.44/0.53 &	0.01/0.42/0.63	& 0.31/0.51/0.47	& 0.04/0.05/0.61	&0.16/0.22/0.51	&0.02/0.3/0.55	& 0.06/0.32/0.25	& 0.0/0.4/0.38	& 0.05/0.1/0.23	& 0.01/0.15/0.36 \\
4\_4              & 0.41/0.47/0.55	& 0.02/0.4/0.65& 	0.22/0.0/0.53& 	0.03/0.0/0.67& 	0.13/0.16/0.49& 	0.01/0.23/0.56	& 0.27/0.33/0.24	& 0.02/0.41/0.37	& 0.24/0.14/0.23	& 0.03/0.19/0.37
\\
4\_7             & 0.27/0.5/0.61	& 0.01/0.44/0.72	& 0.07/0.0/0.45 &	0.02/0.0/0.6	&0.05/0.2/0.49	&0.01/0.25/0.54	&0.06/0.26/0.22	& 0.01/0.31/0.35	&0.17/0.18/0.21	& 0.03/0.25/0.34
\\
4\_11             & 0.16/0.59/0.56	& 0.01/0.54/0.54	& 0.12/0.0/0.3	& 0.02/0.0/0.42	& 0.05/0.56/0.5& 	0.01/0.54/0.55& 	0.35/0.37/0.24& 	0.03/0.43/0.37& 	0.32/0.08/0.21& 	0.04/0.14/0.35

\\
4\_17             & 0.0/0.0/0.57	&0.0/0.0/0.59&	0.43/0.0/0.35	&0.06/0.0/0.51	&0.02/0.14/0.62	&0.01/0.22/0.59	&0.37/0.0/0.23	&0.05/0.0/0.34&	0.17/0.0/0.19	&0.02/0.0/0.31
\\
4\_20             & 0.3/0.34/0.29	 &0.02/0.46/0.36	 &0.27/0.57/0.33	 &0.03/0.31/0.43	 &0.18/0.21/0.25	 &0.03/0.29/0.28	 &0.0/0.0/0.23	 &0.0/0.0/0.36	 &0.22/0.22/0.21 &0.03/0.35/0.35
\\
4\_21             & 0.51/0.0/0.47	 &0.06/0.0/0.52	 &0.27/0.0/0.5	 &0.06/1/0.57	 &0.02/0.0/0.36	 &0.01/0.0/0.4	 &0.0/0.0/0.25	 &0.0/0.0/0.38 &	0.43/0.0/0.2 &	0.06/1/0.32
\\
4\_22             & 0.0/0.0/0.47	 &0.0/0.0/0.39	 &0.17/0.0/0.31	 &0.04/1/0.23 &	0.0/0.0/0.33	 &0.0/0.0/0.31	 &0.25/0.0/0.22	 &0.05/0.0/0.3	 &0.14/0.0/0.17	 &0.04/1/0.26
\\
11\_2             & 0.48/0.41/0.46	 &0.03/0.56/0.46	 &0.18/0.42/0.46	 &0.04/0.53/0.57	 & 0.15/0.34/0.32	 & 0.03/0.47/0.34	 & 0.14/0.24/0.14	 & 0.01/0.38/0.21	 & 0.05/0.16/0.1 &	0.01/0.28/0.16
\\
11\_3             & 0.27/0.41/0.46	 &0.01/0.49/0.5	 &0.47/0.44/0.41	 &0.07/0.59/0.53	 & 0.21/0.26/0.33 &	0.04/0.37/0.37	 &0.1/0.25/0.14	 &0.01/0.4/0.22	 &0.18/0.15/0.1	 &0.02/0.26/0.16
\\
11\_4             & 0.43/0.44/0.42	 &0.02/0.57/0.44	 &0.36/0.4/0.43	 &0.04/0.5/0.55	 & 0.19/0.64/0.33	 & 0.03/0.66/0.33	 & 0.08/0.23/0.16	 & 0.01/0.36/0.24	 & 0.05/0.18/0.1	 & 0.01/0.3/0.17
\\
11\_5             & 0.37/0.39/0.41	 &0.02/0.49/0.43	 &0.16/0.45/0.42	 &0.03/0.55/0.54	 &0.18/0.7/0.33	 &0.04/0.72/0.34	 &0.07/0.26/0.17	 &0.01/0.39/0.25	 &0.25/0.18/0.1	 &0.01/0.3/0.16
\\
11\_6             & 0.22/0.38/0.43	 &0.01/0.46/0.47	 &0.14/0.52/0.39	 &0.02/0.53/0.52	 &0.16/0.68/0.33	 &0.05/0.7/0.34	 &0.02/0.24/0.14	 &0.0/0.37/0.21	 &0.0/0.17/0.12	 &0.0/0.27/0.18
\\
11\_7             & 0.41/0.38/0.47	 &0.03/0.36/0.51	 &0.16/0.34/0.48	 &0.04/0.39/0.62	 &0.12/0.55/0.35	 &0.01/0.58/0.37	 &0.04/0.23/0.18	 &0.01/0.37/0.27	 &0.0/0.16/0.12	 &0.0/0.26/0.2
\\
11\_8             & 0.4/0.4/0.4	 & 0.01/0.39/0.41	 &0.13/0.33/0.37	 &0.04/0.43/0.46	 &0.18/0.66/0.31	 &0.01/0.68/0.31 &	0.14/0.23/0.14	 &0.02/0.36/0.21 &	0.05/0.17/0.09	 &0.01/0.27/0.14
\\
11\_9             & 0.0/0.6/0.32	 &0.0/0.45/0.34	 &0.12/0.4/0.37  &	0.03/0.55/0.46	 &0.01/0.7/0.25	 &0.0/0.72/0.26 &	0.0/0.24/0.13	 &0.0/0.38/0.2	 &0.0/0.19/0.11	 &0.0/0.32/0.17
\\ 
\hline
{\bf{Mean}}    & {\bf{0.27}}/{\underline{0.41}}/{\bf{0.46}}     & {{0.02}}/\underline{0.43}/\underline{0.50}    
& {\underline{0.22}}/0.31/{\underline{0.41}}    & {\bf{0.04}}/0.32/{\bf{0.52}}    
& 0.11/{\bf{0.43}}/0.38    & 0.02/{\bf{0.48}}/{{0.40}}  
& 0.12/0.23/0.19      & 0.01/0.33/0.29     
& 0.15/0.15/0.16      & {{0.02}}/{{0.24}}/0.25 
\\
\hline
\hline
\makecell{\bf{Overall} \\ {\bf{mean}}}    & \multicolumn{1}{c}{\bf{0.38}}     & \multicolumn{1}{c|}{\bf 0.32}   
& \multicolumn{1}{c}{\underline{0.31}}    & \multicolumn{1}{c|}{\underline{0.30}}   
& \multicolumn{1}{c}{\underline{0.31}}    & \multicolumn{1}{c|}{\underline{0.30}}   
& \multicolumn{1}{c}{0.18}      & \multicolumn{1}{c|}{0.21}    
& \multicolumn{1}{c}{0.15}      & \multicolumn{1}{c}{0.17} 
\\
\hline
\hline
\end{tabular}
\end{adjustbox}
\label{table:results_segmentation_2D}
\vspace{5mm}
\caption{\textbf{2D segmentation of landmarks using distance-based metric:} The symmetric distance score $G$ is provided for 16 images from the 2 test cases (patients 4 and 11). Each evaluation metric includes values for the ridge, the falciform ligament, and the silhouette landmarks. The lower the symmetric distance score, the better. The best results are in bold, and the second best are underlined. Mean values for each landmark $\bar{G}$ and for the combined overall mean for all landmarks $\bar{G}_{rls}$ are also provided.}
\begin{adjustbox}{width=0.65\columnwidth,center}
\begin{tabular}{c|c|c|c|c|c}
\hline
\multirow{1}{*}{\textbf{Test}} & \multicolumn{1}{c|}{\multirow{2}{*}{\textbf{BHL}}}                                                        & \multicolumn{1}{c|}{\multirow{2}{*}{\textbf{NCT}}}                                                        & \multicolumn{1}{c|}{\multirow{2}{*}{\textbf{UCL}}}                                                        & \multicolumn{1}{c|}{\multirow{2}{*}{\textbf{VIP}}}                                                        & \multicolumn{1}{c}{\multirow{2}{*}{\textbf{VOR}}}                                                        \\
{\textbf{samples}}    & & & & & \\
\cline{1-6} 
4\_3              &0.67/0.5/0.14	&0.33/0.87/0.3&	0.61/1.0/0.21&	0.77/0.51/0.71&	0.65/1.0/0.87        \\
4\_4              & 0.56/0.6/0.14	&0.65/1.0/0.08	&0.62/1.0/0.34&	0.57/0.47/0.73	&0.43/1.0/0.79      \\
4\_7              & 0.77/0.45/0.08	&0.69/0.84/0.35&	0.62/1.0/0.22&	0.74/0.71/0.9	&0.67/1.0/1.0 \\
4\_11              &0.76/0.37/0.34&	0.7/1.0/0.99	&0.72/0.28/0.18	&0.58/0.44/0.74	&0.59/1.0/1.0 \\
4\_17              & 1.0/1.0/0.35	&0.4/1.0/0.85&	1.0/1.0/0.37	&0.44/1.0/0.89	&0.68/1.0/1.0   \\
4\_20              &0.48/0.77/0.39&	0.58/0.43/0.26	&0.61/1.0/0.4&	1.0/1.0/1.0&	0.54/0.88/1.0    \\
4\_21           & 0.36/NA/0.34	&0.34/NA/0.33&	0.81/NA/0.45&	1.0/NA/0.75	&0.39/NA/1.0  \\
4\_22              & 1.0/NA/0.57&	0.4/NA/0.81	&1.0/NA/0.53&	0.28/NA/1.0&	0.29/NA/1.0    \\
11\_2              & 0.5/0.19/0.5&	0.43/0.22/0.31&	0.53/1.0/0.63&	0.59/0.66/1.0	&0.76/1.0/1.0     \\
11\_3              &0.63/0.19/0.41&	0.37/0.16/0.24&	0.46/1.0/0.49	&0.58/0.65/1.0&	0.61/1.0/1.0  \\
11\_4              & 0.63/0.13/0.36&	0.44/0.27/0.26&	0.6/0.07/0.46&	0.68/0.73/1.0	&0.85/1.0/1.0    \\
11\_5             & 0.59/0.3/0.38&	0.53/0.27/0.32&	0.55/0.05/0.46&	0.7/0.6/1.0	&0.79/1/1  \\
11\_6             & 0.79/0.3/0.38&	0.54/0.39/0.51&	0.61/0.04/0.47&	0.81/0.8/1.0	&1.0/1.0/1.0 \\
11\_7              &0.65/0.58/0.27&	0.56/0.46/0.13&	0.79/0.11/0.32	&0.84/1.0/1.0&	1.0/1.0/1.0  \\
11\_8             & 0.81/0.56/0.5&	0.63/0.36/0.46&	0.81/0.05/0.57	&0.65/1.0/1.0	&0.76/1/1  \\
11\_9             & 0.82/0.53/0.51	&0.46/0.26/0.45&	1.0/0.06/0.57&	1.0/0.79/1.0&	1.0/1.0/1.0 \\
\hline
$\bar{G}$    & \underline{0.69}/\bf{0.46}/\bf{0.35 }     & {\bf{0.50}}/\underline{0.54}/\underline{0.42}      & 0.71/0.55/\underline{0.42}      & 0.70/0.74/0.92       & \underline{0.69}/0.99/0.98      \\
\hline
\hline
\makecell{ $\bar{G}_{rls}$}    & \multicolumn{1}{c|}{\underline{0.50}}    & \multicolumn{1}{c|}{{\bf 0.49}}       & \multicolumn{1}{c|}{0.56}    & \multicolumn{1}{c|}{0.79}      & \multicolumn{1}{c}{0.87}  
\\
\hline
\hline
\end{tabular}
\end{adjustbox}
\label{table:results_segmentation_2D_symmetric}
\end{table}
\end{landscape}
\begin{table*}
\centering
\caption{\textbf{Segmentation of 3D landmarks:} 3D Chamfer distances for the ridge `r' (\textbf{ch\_r}), and the falciform ligament `l' (\textbf{ch\_l}) landmarks are provided for 16 images of the 2 test cases (patients 4 and 11). The ground-truth 3D vertex locations are compared with the predicted 3D vertex locations for the ridge and the falciform ligament. F refers to the failure cases. The best results are in bold, the second best are underlined. Mean values are computed for all the images except for the failed cases shown in red. The overall mean is computed as an average of ch\_r and ch\_l for
each team. All the distances are given in millimeters.}
\begin{adjustbox}{width=1.5\columnwidth,center}
\begin{tabular}{c|cc|cc|cc|cc}
\hline
\multirow{1}{*}{\textbf{Test}} & \multicolumn{2}{c|}{\textbf{BHL}}                      & \multicolumn{2}{c|}{\textbf{NCT}}                      & \multicolumn{2}{c|}{\textbf{UCL}}                      & \multicolumn{2}{c}{\textbf{VOR}}                      \\ \cline{2-9} 
                   {\textbf{\#1}}            & \multicolumn{1}{l}{\textbf{ch\_r}} & \textbf{ch\_l} & \multicolumn{1}{l}{\textbf{ch\_r}} & \textbf{ch\_l} & \multicolumn{1}{l}{\textbf{ch\_r}} & \textbf{ch\_l} & \multicolumn{1}{l}{\textbf{ch\_r}} & \textbf{ch\_l} \\ \hline
4\_3                        & 98.02                         & 69.98                     & 20.14                        & 37.95                     & 7.68                         & 14.45                     & 28.86                        & 22.88                     \\ \hline
4\_4                        & 102.83                        & 61.08                     & 17.82                        & 38.96                     & 16.69                        & 10.92                     & 33.41                        & 24.19        \\ \hline
4\_7                        & 84.99                         & 55.15                     & 24.59                        & 39.3                      & 10.2                         & 11.95                     & 26.23                        & 20.82   \\ \hline
4\_11                       & 101.65                        & 52.19                     & 19.46                        & 40.27                     & 15.07                        & 17.21                     & 34.12                        & 22.35      \\ \hline
4\_17                       & 105.7                         & 78.52                     & 20.76                        & 37.77                     & 7.61                         & 16.43                     & 34.21                        & 30.46             \\ \hline
4\_20                       & 108.05                        & 78.91                     & 20.55                        & 39.68                     & 17.42                        & 11.91                     & 31.62                        & 18.46   \\ \hline
\rowcolor[HTML]{C0C0C0}{\color[HTML]{FF0000}4\_21}                       &
{\color[HTML]{FF0000} 156.74} & {\color[HTML]{FF0000} F} & {\color[HTML]{FF0000} 36.83} & {\color[HTML]{FF0000} F} & {\color[HTML]{FF0000} 13.17} & {\color[HTML]{FF0000} F} & {\color[HTML]{FF0000} 40.39} & {\color[HTML]{FF0000} F} \\   \hline
\rowcolor[HTML]{C0C0C0}{\color[HTML]{FF0000}4\_22}                     &
{\color[HTML]{FF0000} 148.92} & {\color[HTML]{FF0000} F} & {\color[HTML]{FF0000} 33.8}  & {\color[HTML]{FF0000} F} & {\color[HTML]{FF0000} 38.05} & {\color[HTML]{FF0000} F} & {\color[HTML]{FF0000} 35.5}  & {\color[HTML]{FF0000} F}              \\ \hline
11\_2                       &146.26                        & 57.72                     & 30.16                        & 33.67                     & 37.49                        & 24.24                     & 29.92                        & 36.62      \\ \hline
11\_3                       & 136.32                        & 63.43                     & 30.63                        & 35.55                     & 45.13                        & 31.49                     & 28.29                        & 33.54            \\ \hline
11\_4                       & 145.55                        & 63.31                     & 30.97                        & 37.04                     & 52.76                        & 34.68                     & 37.42                        & 33.88             \\ \hline
11\_5                       & 152.12                        & 57.72                     & 31.61                        & 32.92                     & 15.28                        & 36.89                     & 32.52                        & 35.13       \\ \hline
11\_6                       & 143.85                        & 57.86                     & 30.77                        & 33.32                     & 14.23                        & 28.07                     & 30.39                        & 37.64          \\ \hline
11\_7                       & 145.3                         & 58.32                     & 32.15                        & 34.95                     & 70.82                        & 43.51                     & 33.19                        & 37.59      \\ \hline
11\_8                       & 152.87                        & 53.45                     & 33.3                         & 34.59                     & 68.01                        & 18.17                     & 37.18                        & 33.31                     \\ \hline
11\_9                       & 172.22                        & 60.76                     & 37.8                         & 33.41                     & 13.23                        & 42.62                     & 43.22                        & 35.15             \\ 
\hline
{\bf{Mean}}                      & 128.27                        & 62.02                     & \bf{27.19}                       & 36.38                     & \underline{27.97}                        & \bf{24.47}                    & 32.90                        & \underline{30.14}             
\\
\hline
 \hline
\makecell{\bf{Overall mean}}    & \multicolumn{2}{c|}{95.14}     & \multicolumn{2}{c|}{{31.78}}    & \multicolumn{2}{c|}{\textbf{26.22}}    & \multicolumn{2}{c}{\underline{31.52}}  
\\
\hline
\hline
\end{tabular}
\label{table:results_segmentation_3D}
\end{adjustbox}
\end{table*}
\begin{table*}
\centering
\caption{\textbf{3D-2D registration:} Reprojection errors in pixels are provided for 16 samples from the two test patients. These errors are computed for the ridge (rpe\_r) and the ligament (rpe\_l) between the projected 3D ground-truth landmark vertices in the registered model w.r.t. the 2D ground-truth pixel locations. F refers to the failure cases. Mean values are computed for all registrations except for the failed cases shown in red. The overall mean is computed as an average of rpe\_r and rpe\_l for each team. The best results are in bold, the second best are underlined.}
\begin{adjustbox}{width=1.6\columnwidth,center}
\begin{tabular}{c|cc|cc|cc|cc|cc}
\hline
\multicolumn{1}{c|}{\multirow{2}{*}{\textbf{Test}}} & \multicolumn{2}{c|}{\textbf{BHL}}                                           & \multicolumn{2}{c|}{\textbf{NCT}}                                               & \multicolumn{2}{c|}{\textbf{UCL}}                                           & \multicolumn{2}{c|}{\textbf{VOR}}                                           & \multicolumn{2}{c}{\textbf{GRASP}}                                         \\ \cline{2-11} 
\multicolumn{1}{c|}{}                                                                              & \multicolumn{1}{l}{\textbf{rpe\_r}} & \multicolumn{1}{l|}{\textbf{rpe\_l}} & \multicolumn{1}{l}{\textbf{rpe\_r}}     & \multicolumn{1}{l|}{\textbf{rpe\_l}} & \multicolumn{1}{l}{\textbf{rpe\_r}} & \multicolumn{1}{l|}{\textbf{rpe\_l}} & \multicolumn{1}{l}{\textbf{rpe\_r}} & \multicolumn{1}{l|}{\textbf{rpe\_l}} & \multicolumn{1}{l}{\textbf{rpe\_r}} & \multicolumn{1}{l}{\textbf{rpe\_l}} \\ \hline
4\_3                                                                                                & \multicolumn{1}{l}{515.89}          & 702.65                               & \multicolumn{1}{l}{401.36}              & 257.95                               & \multicolumn{1}{l}{525.82}          & 708.6                                & \multicolumn{1}{l}{936.33}          & 499.04                               & \multicolumn{1}{l}{446.93}          & 661.98                               \\ \hline
4\_4                                                                                                & \multicolumn{1}{l}{744.36}          & 1050.4                               & \multicolumn{1}{l}{494.53}              & 368.75                               & \multicolumn{1}{l}{732.54}          & 466.69                               & \multicolumn{1}{l}{1035.62}         & 567.85                               & \multicolumn{1}{l}{521.31}          & 762.17                               \\ \hline
4\_7                                                                                                & \multicolumn{1}{l}{431.06}          & 398.92                               & \multicolumn{1}{l}{115.73}              & 170.76                               & \multicolumn{1}{l}{656.34}          & 366.39                               & \multicolumn{1}{l}{869.04}          & 480.69                               & \multicolumn{1}{l}{474.44}          & 558.58                               \\ \hline
4\_11                                                                                               & \multicolumn{1}{l}{857.2}           & 901.19                               & \multicolumn{1}{l}{360.19}              & 329.4                                & \multicolumn{1}{l}{340.81}          & 479.25                               & \multicolumn{1}{l}{979.86}          & 669.47                               & \multicolumn{1}{l}{443.22}          & 682.49                               \\ \hline
4\_17                                                                                               & \multicolumn{1}{l}{500.09}          & 3182.76                              & \multicolumn{1}{l}{323.6}               & 458.22                               & \multicolumn{1}{l}{250.71}          & 442.19                               & \multicolumn{1}{l}{1142.3}          & 853.63                               & \multicolumn{1}{l}{405.36}          & 444.67                               \\ \hline
4\_20                                                                                               & \multicolumn{1}{l}{664.21}          & 946.64                               & \multicolumn{1}{l}{183.58}              & 393.21                               & \multicolumn{1}{l}{707.26}          & 300.48                               & \multicolumn{1}{l}{992.22}          & 762                                  & \multicolumn{1}{l}{495.07}          & 652.99                               \\ \hline
\rowcolor[HTML]{C0C0C0}{\color[HTML]{FF0000}4\_21}                                                                                               & \multicolumn{1}{l}{{\color[HTML]{FF0000}448.63}}          & {\color[HTML]{FF0000}F}                                    & \multicolumn{1}{l}{{\color[HTML]{FF0000}159.3}}               & {\color[HTML]{FF0000}F}                                    & \multicolumn{1}{l}{{\color[HTML]{FF0000}799.6}}           & {\color[HTML]{FF0000}F}                                    & \multicolumn{1}{l}{{\color[HTML]{FF0000}976.17}}          & {\color[HTML]{FF0000}F}                                    & \multicolumn{1}{l}{{\color[HTML]{FF0000}451.3}}           & {\color[HTML]{FF0000}F}                                    \\ \hline
4\_22                                                                                               & \multicolumn{1}{l}{465.4}           & 850.12                               & \multicolumn{1}{l}{212.36}              & 209.94                               & \multicolumn{1}{l}{604.88}          & 291.51                               & \multicolumn{1}{l}{965.59}          & 973.99                               & \multicolumn{1}{l}{504.12}          & 333.38                               \\ \hline
11\_2                                                                                               & \multicolumn{1}{l}{839.91}          & 2019.02                              & \multicolumn{1}{l}{1008.61}             & 356.36                               & \multicolumn{1}{l}{910.76}          & 473.38                               & \multicolumn{1}{l}{1293.85}         & 1194.77                              & \multicolumn{1}{l}{873.09}          & 452.32                               \\ \hline
11\_3                                                                                               & \multicolumn{1}{l}{520.78}          & 3115.51                              & \multicolumn{1}{l}{842.67}              & 177.02                               & \multicolumn{1}{l}{613.39}          & 492.87                               & \multicolumn{1}{l}{1253.12}         & 1170.4                               & \multicolumn{1}{l}{794.59}          & 473.93                               \\ \hline
11\_4                                                                                               & \multicolumn{1}{l}{508}             & 659.02                               & \multicolumn{1}{l}{720.35}              & 185.74                               & \multicolumn{1}{l}{974.13}          & 697.18                               & \multicolumn{1}{l}{1278.17}         & 574.81                               & \multicolumn{1}{l}{792.05}          & 542.79                               \\ \hline
11\_5                                                                                               & \multicolumn{1}{l}{403}             & 688.8                                & \multicolumn{1}{l}{788.44}              & 311.52                               & \multicolumn{1}{l}{825.24}          & 669.36                               & \multicolumn{1}{l}{1281.08}         & 583.92                               & \multicolumn{1}{l}{767.99}          & 608.84                               \\ \hline
11\_6                                                                                               & \multicolumn{1}{l}{509.56}          & 710.12                               & \multicolumn{1}{l}{807.11}              & 543.89                               & \multicolumn{1}{l}{936.63}          & 662.15                               & \multicolumn{1}{l}{1283.26}         & 531.75                               & \multicolumn{1}{l}{1107.94}         & 463.24                               \\ \hline
11\_7                                                                                               & \multicolumn{1}{l}{489.7}           & 670.1                                & \multicolumn{1}{l}{360.03}              & 408.01                               & \multicolumn{1}{l}{1058.9}          & 816.27                               & \multicolumn{1}{l}{1248.58}         & 511.68                               & \multicolumn{1}{l}{1291.6}          & 1587.21                              \\ \hline
11\_8                                                                                               & \multicolumn{1}{l}{388.29}          & 522.38                               & \multicolumn{1}{l}{329.8}               & 237.32                               & \multicolumn{1}{l}{1174.43}         & 824.16                               & \multicolumn{1}{l}{1279.16}         & 472.45                               & \multicolumn{1}{l}{781.51}          & 627.21                               \\ \hline
11\_9                                                                                               & \multicolumn{1}{l}{247.95}          & 369.42                               & \multicolumn{1}{l}{361.25}              & 270.65                               & \multicolumn{1}{l}{925.53}          & 679.32                               & \multicolumn{1}{l}{1250.07}         & 746.84                               & \multicolumn{1}{l}{753.88}          & 643.11                               \\ \hline
\bf{Mean}  & \multicolumn{1}{l}{533.38}      & 1119.14                          & \multicolumn{1}{l}{466.80} & 311.91                     & \multicolumn{1}{l}{752.31}      & {557.98}                       & \multicolumn{1}{l}{1129.02}      & 706.22                         & \multicolumn{1}{l}{681.52}         & 632.99                              \\
\hline
\hline
\makecell{\bf{Overall mean}}    & \multicolumn{2}{c|}{826.26}     & \multicolumn{2}{c|}{\textbf{389.35}}    & \multicolumn{2}{c|}{\underline{655.14}}    & \multicolumn{2}{c|}{917.62}    & \multicolumn{2}{c}{657.25}
\\
\hline
\hline
\end{tabular}
\label{table:results_registration}
\end{adjustbox}
\end{table*}
\section*{Quantitative results}
\label{sec:results_quantitative}
\subsection*{Performance comparison for the 2D and 3D segmentation task}
%

For the 2D segmentation {step}, the quantitative results for the precision and the Dice coefficient (DSC) scores are presented in Table \ref{table:results_segmentation_2D}. The {GRASP} team did not perform detection of the anatomical landmarks. In terms of precision, the BHL team has the best overall mean precision, with $0.38$, and the best mean scores for the ridge and silhouette landmarks, with $0.27$ and $0.46$, respectively. They also have the second best mean score for the ligament landmark, with $0.41$. The second best results are for the NCT and UCL teams, with an overall mean precision of $0.31$. The NCT team has the second best scores for the ridge and silhouette landmarks, with $0.22$ and $0.41$, respectively. The UCL team has the best score for the ridge landmark, with $0.43$. In terms of DSC, the BHL team has the best overall mean score, with $0.32$. They also have the second best scores for the ligament and silhouette landmarks, with $0.43$ and $0.50$, respectively.~The NCT and UCL teams have both the second best overall mean score, with $0.30$. The NCT team has the best scores for the ridge and silhouette landmarks, with $0.04$ and $0.52$, respectively. The UCL team has the best score for the ligament landmark, with $0.48$. The VIP and VOR team performed poorly on both metrics.
Contrary to the precision and the DSC, the lower symmetric distance score, the better. These results are shown in Table \ref{table:results_segmentation_2D_symmetric}. The NCT team has the best overall mean symmetric distance score, with $0.49$. They have the best mean score for the ridge landmark, with $0.50$, and the second best mean scores for the ligament and silhouette landmarks, with $0.54$ and $0.42$, respectively. The BHL team has the second best overall mean symmetric distance score, with $0.50$. They have the best mean scores for the ligament and silhouette landmarks, with $0.46$ and $0.35$, respectively. They also have the second best score for the ridge landmark along with the VOR team, with $0.69$. In general, we can observe that the ligament and the silhouette landmarks have the best prediction performances, followed by the ridge landmark which has the worst segmentation performance by all teams.\\
\indent{For} the 3D segmentation {step}, the quantitative results are presented in Table \ref{table:results_segmentation_3D}. The UCL team had the best overall scores, with the lowest distance for the ligament landmark at $24.47$ mm and the second lowest distance for the ridge landmark at $27.97$ mm. The NCT had the lowest distance for the ridge landmark at $27.19$ mm. The VOR team had the second lowest distance for the ligament landmark at $30.14$ mm.

\subsection*{Performance comparison for the 3D-2D registration task}
The quantitative results for the five teams participating in the 3D-2D registration task are presented in Table \ref{table:results_registration}. Reprojection errors are computed for the ridge and ligament landmarks. We analyse the mean reprojection errors of both landmarks as doing it separately does not provide enough information on the accuracy of the registered models. The VIP team did not participate in this task. The team NCT has the best mean reprojection error with $389.35$ px. The team UCL has the second best mean error, with $655.14$ px. Following closely, the {GRASP} team has the third best mean error with $657.25$ px.
\section*{Qualitative results}
\label{sec:results_qualitative}
We present a visual representation of the results obtained by the teams in both tasks. For Task 1, we show a side-by-side comparison of the predicted and the ground-truth 2D and 3D landmarks. For Task 2, we overlay the registered 3D models on top of the laparoscopic images of the two test patients.

\subsection*{Segmentation of 2D landmarks}
Figure \ref{fig:qualitative_results_landmarks_2D} shows the ground-truth and predicted 2D landmarks for three images of the two test patients. The images correspond, from the left most column to the right most column, to the cases 4\_7, 4\_11, 4\_17, 11\_3, 11\_6, and 11\_7 of Table \ref{table:results_segmentation_2D}. In general, all the teams were able to detect the ridge, ligament, and silhouette landmarks in the laparoscopic images. Visually speaking, the quality of the predictions correspond to the scores reported in Table \ref{table:results_segmentation_2D}, with the BHL and NCT teams having less spurious predictions compared to the other teams. The silhouette landmark is the one with the best predictions across all the teams, with more continuous curves and less missing parts. The ligament landmark also has good results with continuous curves and low spurious responses. The ridge landmark is the most challenging landmark, with lots of missing parts and a considerable amount of spurious predictions.

\subsection*{Segmentation of 3D landmarks}
Figure \ref{fig:qualitative_results_landmarks_3D} shows the ground-truth and predicted 3D landmarks for the same set of images presented in figure \ref{table:results_segmentation_2D}. The landmarks shown correspond to the ridge and the falciform ligament. The BHL team was not able to clearly segment the landmarks, segmenting vertices that are far from the ground-truth locations and covering large areas of the liver surface. The NCT team was able to detect the ridge landmarks successfully, while the ligament landmarks present some spurious responses. The team has segmented the landmarks in the whole 3D model, rather than in a per-image basis, which was the original goal of the task. The UCL team has detected the ridge landmarks successfully for patient 4, with some spurious responses for the ligament landmarks. For patient 11, the ridge landmarks are not consistent and the ligament landmarks are not clearly defined. Although the team segments the landmarks in the whole 3D model and not on a per-view basis, their method gives different responses when run multiple times on the same model. The VOR team was not able to succesfully segment the ridge landmarks, while the ligament landmarks are far from the ground-truth ones and present some spurious responses. The team has also segmented the landmarks in the whole liver, having the same responses at every running instance.

\subsection*{3D-2D registration}
Figure \ref{fig:qualitative_results_registration} shows a fusion of the registered 3D models with the laparoscopic images of figure \ref{table:results_segmentation_2D}. Matching the results of table \ref{table:results_registration}, the NCT team has the best visual results, with the registered models having a similar pose to the intraoperative livers. However, the models do not exactly fit the boundaries of the intraoperative livers, which means that using them for AR purposes would be inaccurate. The rest of the methods did not provide visually successful results, with the registered models having different poses or being far from the intraoperative livers. Results for the VOR team are not shown due to their registered models falling out of the laparoscopic images.

\section*{Discussion}
\label{sec:discussion}
%
%
%
\begin{figure*}
    \centering
    \includegraphics[width=0.92\textwidth]{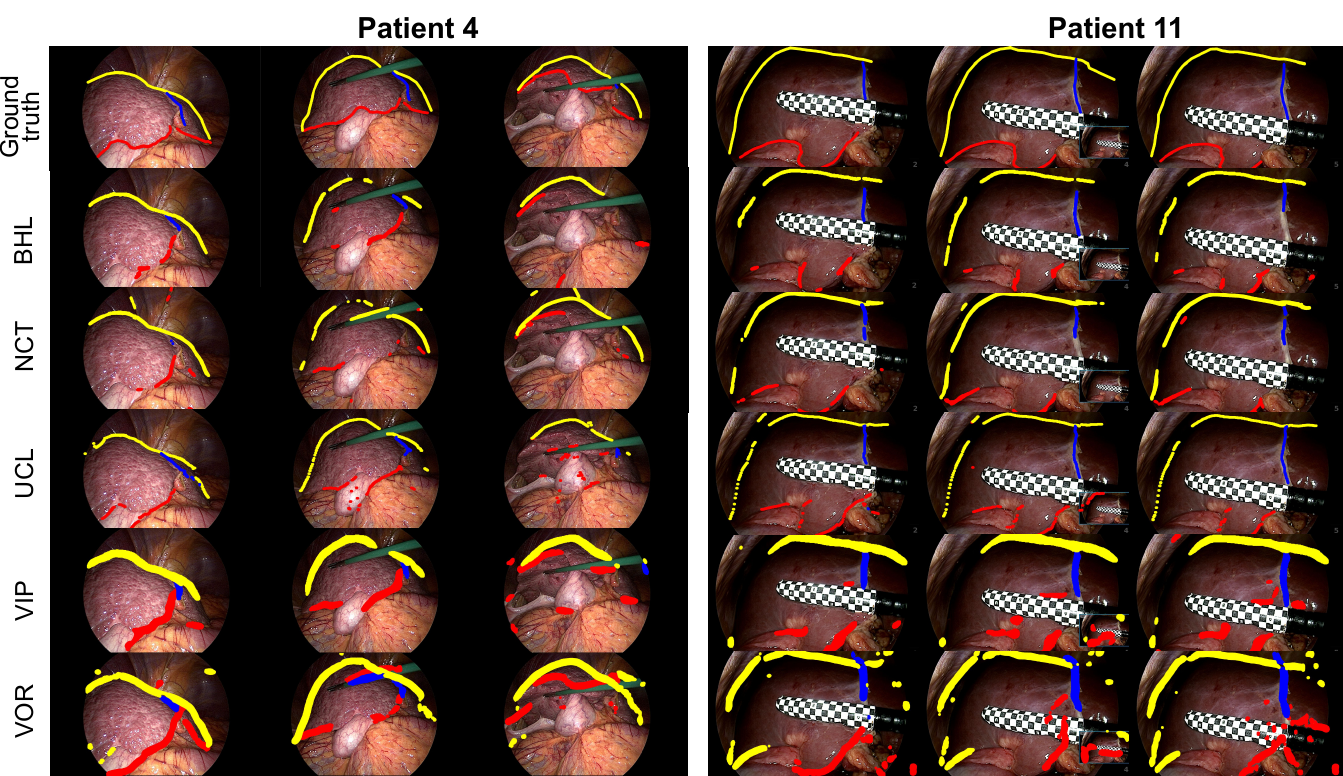}
    \caption{\textbf{Qualitative results of the 2D landmark segmentation task:} The ground-truth (GT) landmarks for the two test patients are shown in the first row, while the teams' predictions are shown in the consecutive rows. The ridge landmarks are shown in red, the ligament landmarks in blue, and the silhouette landmarks in yellow.}
    \label{fig:qualitative_results_landmarks_2D}
\end{figure*}
\begin{figure*}
    \centering
    \includegraphics[width=0.92\textwidth]{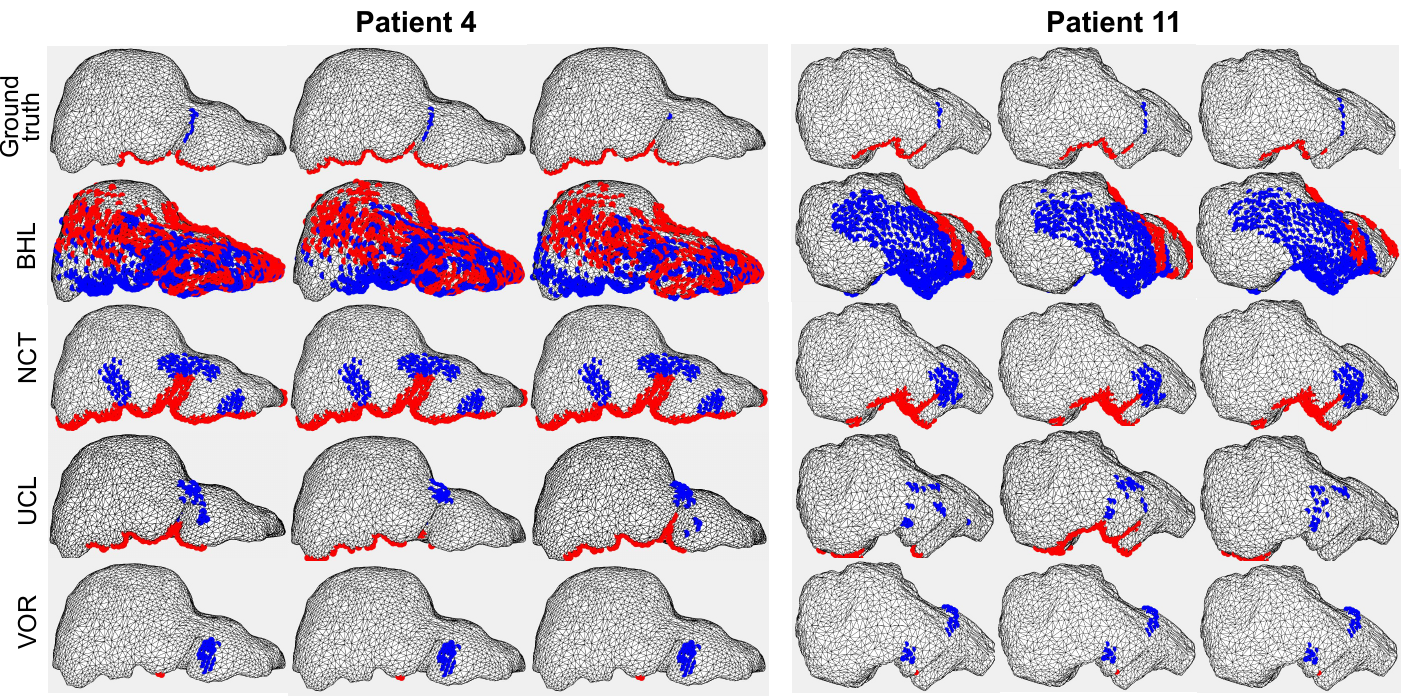}
    \caption{\textbf{Qualitative results of the 3D landmark segmentation task:} The ground-truth (GT) landmarks for the two test patients are shown in the first row, while the teams' predictions are shown in the consecutive rows. The ridge landmarks are shown in red and the ligament landmarks in blue.}
    \label{fig:qualitative_results_landmarks_3D}
\end{figure*}
\begin{figure*}
    \centering
    \includegraphics[width=\textwidth]{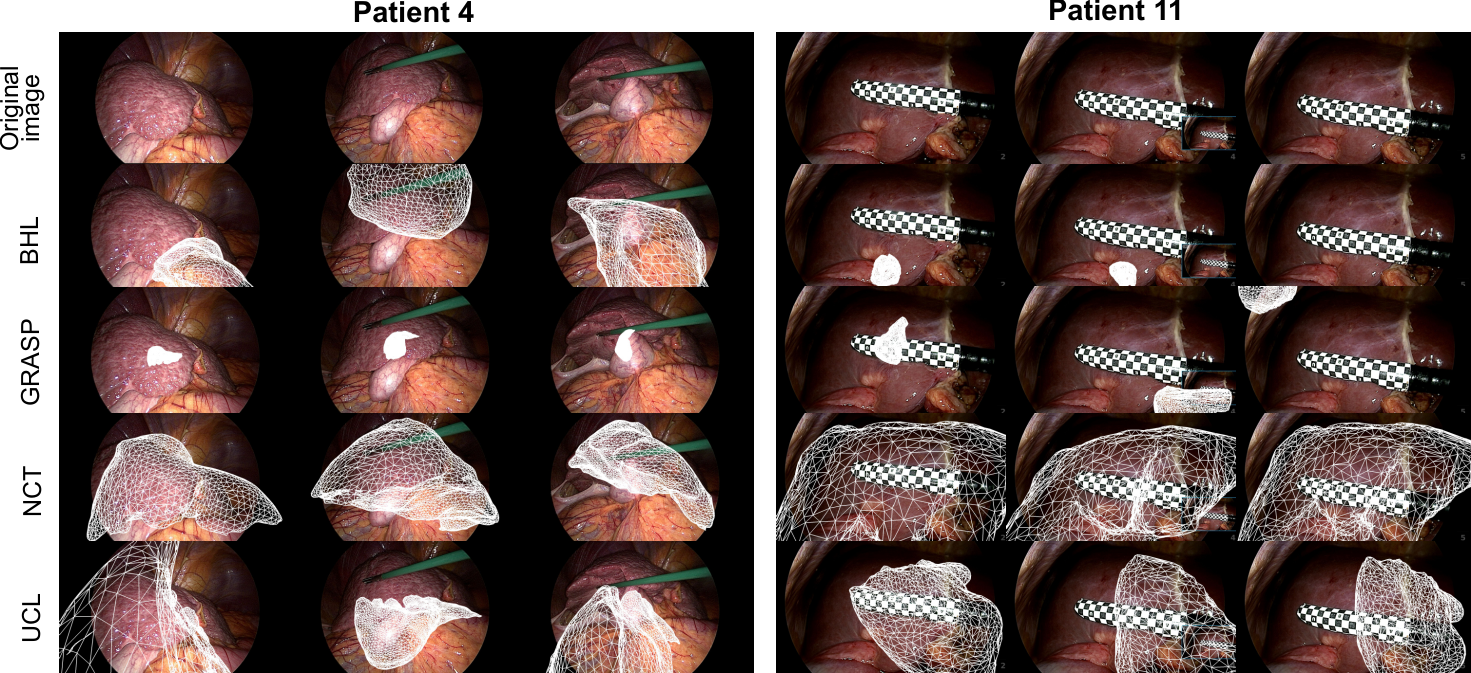}
    \caption{\textbf{Qualitative results of the 3D-2D registration task:} {Registration results on some of the images are shown for 4 of the participating teams. The original images are shown in the first row, with the resutls for BHL, GRASP, NCT, and UCL shown in the consecutive rows. Results for VOR are not shown due to the models being out of the field of view. It can be seen that NCT obtained the best results, with the registered models being close to the liver in the images.}}
    \label{fig:qualitative_results_registration}
\end{figure*}
Evident from the literature, there is a clear lack of publicly available datasets, and most of the existing AR methods use non-publicly available data and deal only with the 2D landmark automatic detection problem. For example, \cite{koo2022} used a private dataset of 133 images coming from two patients to detect the anatomical landmarks of the liver ({including} silhouette and ridge). The method heavily relied on synthetic data generation.~\cite{labrunie2022} used a dataset of 1415 laparoscopic images coming from 68 patients for detecting the liver landmarks, but such dataset is not publicly available either. There are some available datasets containing endoscopic liver videos like the Cholec80 dataset~\citep{twinanda2016}, HeiChole dataset~\citep{wagner2023}, and the Dresden Surgical Anatomy Dataset~\citep{carstens2023}, but they do not contain the preoperative data and the intrinsic camera parameters required to work with AR.
Through this challenge, we aim to release the first comprehensive dataset with carefully annotated anatomical landmarks in both the laparoscopic images and the preoperative 3D models. The 2D landmarks consist of the silhouette, ridge and falciform ligament (often persisted as a potential anatomical landmark by surgeons), while the 3D landmarks consist of the falciform ligament and the ridge. An important limitation of the preoperative to intraoperative registration problem is the validation of techniques, as there is no standard validation strategy. We argue that measuring the reprojection error between 2D and 3D landmarks is a valid strategy, although it carries an ambiguity problem as different registrations can lead to similar reprojection errors. Therefore, it does not fully replace a proper target registration error measurement using reliable 3D landmarks.

Most of the participating teams developed methods for both the landmark segmentation and the 3D-2D registration tasks. In terms of 2D landmark segmentation, the majority of teams explored various encoder-decoder UNet-based variants (e.g, Attention UNet, UNet++, nnUNet) either alone or in combination with other models and backbones. In the reports submitted to the challenge organisers, the teams mentioned that using the UNet architecture alone was not enough for having good predictions. The NCT team used nnUNet with dilation as preprocessing, which encouraged false positives (i.e., penalising precision and resulting in the second best score with 0.31 compared to 0.38 from BHL). However, they did obtain the lowest symmetric distance score between their predictions and the ground-truth landmarks with 0.49. It is to be noted that the team did data augmentation and used five-fold cross validation, unifying the results from the five trained models. Similarly, the UCL team used UNet++ with a heavy data augmentation, generating 100,000 synthetic images for training. The BHL team used two ResUNet, one to predict the ridge from the original images, and the other to predict the ligament and the silhouette from contour-enhanced images. In this way, the team achieved the best precision in all the three landmarks. 
\textit{Hypothesis I: Based on the experimental results, segmenting anatomical landmarks in the liver is extremely challenging. Our exploration concluded that using complex model designs can provide a higher precision, as shown by the BHL team. Using synthetic data for training can improve performance, as shown by the UCL team.}

In the context of the segmentation of 3D anatomical landmarks, all the teams performed a global 3D landmark segmentation, instead of a per-view one. Two of the teams (BHL and UCL) utilised PointNet++, the NCT team used MeshCNN, and the VOR team used a graph CNN-based approach. However, since the ratio of number of annotated vertices to the total number of vertices is very small, teams using off-the-shelf methods without much modification did not succeed in achieving acceptable results (e.g., BHL and VOR). On the other hand, team UCL not only used spectral augmentation during training to simulate nearly 1800 3D liver models, but also a combined Hausdorff distance with a negative log likelihood loss function. Similarly, team NCT utilised two MeshCNN and used synthetic meshes during training. First, they merged both the ridge and ligament landmarks of all the views for each patient. Then, they generated 8208 annotated synthetic meshes. Finally, they merged the results from the two MeshCNN to obtain the final result. It can be observed from Table~\ref{table:results_segmentation_3D} that team UCL achieved the best metrics for the ligament landmark, and second best for the ridge landmark. The NCT team had the most accurate result for the ridge landmark. This is also evident in the qualitative results shown in Figure~\ref{fig:qualitative_results_landmarks_3D}.
\textit{Hypothesis II: From the experimental results, it can be concluded that the segmentation of 3D landmarks requires data augmentation to tackle the class imbalance problem. Also, fusing the landmarks from all the views to obtain global ridge and ligament landmarks might help to improve the segmentation performance.}
%

With respect to the 3D-2D registration problem, most of the teams used differentiable rendering as a way to optimise the liver pose. The main difference between them was the registration constraints used during the optimisation. Among these methods, team NCT obtained the best results both quantitatively and qualitatively. Evidently, it was the only team that had visually satisfactory results. This can be explained by the initialisation step, in which the preoperative 3D model is set at a random location in front of the camera, with a rotation such that the liver's anterior side faces the camera. The BHL team was the only one to use a PnP-based approach to perform registration. Apart from the NCT team, the 3D poses obtained by the other teams were far from the liver in the laparoscopic images. \textit{Hypothesis III: From the methods used by the participating teams, it can be deduced that a good initialisation is required to obtain a successful result. This means that the optimisation should start with a pose of the preoperative 3D model close to the one of the liver in the image. Otherwise, the methods will converge into a wrong result. Similarly, given the fact that the proposed methods only performed rigid registration, it can be concluded that the methods are not ready for usage in AR, as the deformations between the preoperative and intraoperative stages are not compensated, which also reduces the registration accuracy.}

\section*{Conclusion and future directions}
\label{sec:conclusion}
The P2ILF challenge is the first challenge that focuses on both the 2D/3D landmark segmentation and the registration problems for AR in laparoscopy. The participating teams understood the importance of the problem, and proposed relevant solutions to it. Although the proposed idea was to segment the 3D landmarks according to the visible 2D landmarks in a laparoscopic image, the teams treated the 2D and 3D segmentation as separate problems. They achieved this by merging the per-image annotated 3D landmarks for all the views of a patient, generating global 3D landmark annotations. Even if it is possible to combine these global 3D landmarks with the visible 2D landmarks for registration, using only the per-view visible 3D landmarks may improve the registration accuracy. Given the acceptable results for the 2D landmark segmentation and the less accurate 3D segmentation, we can conclude that the 3D segmentation problem is more complex than the 2D segmentation one {and requires deeper research}. This can be due to the small number of 3D models, 9 for the training set, compared to the number of laparoscopic images, 167 for the training set. Regarding the registration task, using differentiable rendering in combination with the predicted landmarks can provide coherent results, given a good initial pose of the preoperative 3D model. However, the preoperative-to-intraoperative deformations should be taken into account for this approach to be used in the operating room. According to these results, a dataset with a larger amount of annotated 2D and 3D data is required to improve landmark segmentation. A better landmark segmentation combined with the preoperative-to-intraoperative deformations should improve 3D-2D registration, which is important to have an accurate AR.
%

%


\paragraph*{Code availability}
To help users with the evaluation of the methods we have shared the evaluation codes used in this manuscript at: 
\url{https://github.com/sharib-vision/P2ILF}.

\paragraph*{Declaration of Competing Interest}
The authors declare that they have no known competing financial interests or personal relationships that could have appeared to influence the work reported in this paper.
%
\paragraph*{Author contributions}
S. Ali, A. Bartoli, Y. Espinel and Y. Jin conceptualised the work, led the challenge and workshop, and prepared the dataset and software. S. Ali and Y. Espinel wrote most of the paper. They did all the analyses with the method contributions from the challenge participants (P. Lue, B. Guttner, X. Zhang, L. Zhang, T. Dowrick, M.J. Clarkson, S. Xiao, Y. Wu, Y. Yang, L. Zhu, D. Sun, L. Li, and M. Pfeiffer) and summarisation assistance from Y. Jin. All authors participated in revising this manuscript, provided input, and agreed to submit it.
%
\bibliographystyle{model2-names.bst}\biboptions{authoryear}
\bibliography{biblio}

\begin{thebibliography}{43}
\expandafter\ifx\csname natexlab\endcsname\relax\def\natexlab#1{#1}\fi
\providecommand{\url}[1]{\texttt{#1}}
\providecommand{\href}[2]{#2}
\providecommand{\path}[1]{#1}
\providecommand{\DOIprefix}{doi:}
\providecommand{\ArXivprefix}{arXiv:}
\providecommand{\URLprefix}{URL: }
\providecommand{\Pubmedprefix}{pmid:}
\providecommand{\doi}[1]{\href{http://dx.doi.org/#1}{\path{#1}}}
\providecommand{\Pubmed}[1]{\href{pmid:#1}{\path{#1}}}
\providecommand{\bibinfo}[2]{#2}
\ifx\xfnm\relax \def\xfnm[#1]{\unskip,\space#1}\fi
\bibitem[{Adagolodjo et~al.(2017)Adagolodjo, Trivisonne, Haouchine, Cotin and
  Courtecuisse}]{adagolodjo2017}
\bibinfo{author}{Adagolodjo, Y.}, \bibinfo{author}{Trivisonne, R.},
  \bibinfo{author}{Haouchine, N.}, \bibinfo{author}{Cotin, S.},
  \bibinfo{author}{Courtecuisse, H.}, \bibinfo{year}{2017}.
\newblock \bibinfo{title}{Silhouette-based pose estimation for deformable
  organs application to surgical augmented reality}, in:
  \bibinfo{booktitle}{2017 IEEE/RSJ International Conference on Intelligent
  Robots and Systems (IROS)}, pp. \bibinfo{pages}{539--544}.
\bibitem[{{Agisoft LLC}(2023)}]{metashape}
\bibinfo{author}{{Agisoft LLC}}, \bibinfo{year}{2023}.
\newblock \bibinfo{title}{{Agisoft Metashape}}.
\newblock \URLprefix \url{https://www.agisoft.com/}.
\bibitem[{Ali et~al.(2022)Ali, Espinel, Jin and Bartoli}]{p2ilf}
\bibinfo{author}{Ali, S.}, \bibinfo{author}{Espinel, Y.}, \bibinfo{author}{Jin,
  Y.}, \bibinfo{author}{Bartoli, A.}, \bibinfo{year}{2022}.
\newblock \bibinfo{title}{{Preoperative to Intraoperative Laparoscopy Fusion
  Challenge (P2ILF) - MICCAI 2022}}.
\newblock \bibinfo{howpublished}{\url{https://p2ilf.grand-challenge.org/}}.
\newblock \bibinfo{note}{[Online; accessed 12-August-2022]}.
\bibitem[{Bernhardt et~al.(2016)Bernhardt, Nicolau, Bartoli, Agnus, Soler and
  Doignon}]{bernhardt2016}
\bibinfo{author}{Bernhardt, S.}, \bibinfo{author}{Nicolau, S.A.},
  \bibinfo{author}{Bartoli, A.}, \bibinfo{author}{Agnus, V.},
  \bibinfo{author}{Soler, L.}, \bibinfo{author}{Doignon, C.},
  \bibinfo{year}{2016}.
\newblock \bibinfo{title}{Using shading to register an intraoperative ct scan
  to a laparoscopic image}, in: \bibinfo{editor}{Luo, X.},
  \bibinfo{editor}{Reichl, T.}, \bibinfo{editor}{Reiter, A.},
  \bibinfo{editor}{Mariottini, G.L.} (Eds.),
  \bibinfo{booktitle}{Computer-Assisted and Robotic Endoscopy}, pp.
  \bibinfo{pages}{59--68}.
\bibitem[{Caron et~al.(2021)Caron, Touvron, Misra, J{\'e}gou, Mairal,
  Bojanowski and Joulin}]{caron2021emerging}
\bibinfo{author}{Caron, M.}, \bibinfo{author}{Touvron, H.},
  \bibinfo{author}{Misra, I.}, \bibinfo{author}{J{\'e}gou, H.},
  \bibinfo{author}{Mairal, J.}, \bibinfo{author}{Bojanowski, P.},
  \bibinfo{author}{Joulin, A.}, \bibinfo{year}{2021}.
\newblock \bibinfo{title}{Emerging properties in self-supervised vision
  transformers}, in: \bibinfo{booktitle}{Proceedings of the IEEE/CVF
  International Conference on Computer Vision}, pp.
  \bibinfo{pages}{9650--9660}.
\bibitem[{Carstens et~al.(2023)Carstens, Rinner, Bodenstedt, Jenke, Weitz,
  Distler, Speidel and Kolbinger}]{carstens2023}
\bibinfo{author}{Carstens, M.}, \bibinfo{author}{Rinner, F.M.},
  \bibinfo{author}{Bodenstedt, S.}, \bibinfo{author}{Jenke, A.C.},
  \bibinfo{author}{Weitz, J.}, \bibinfo{author}{Distler, M.},
  \bibinfo{author}{Speidel, S.}, \bibinfo{author}{Kolbinger, F.R.},
  \bibinfo{year}{2023}.
\newblock \bibinfo{title}{The dresden surgical anatomy dataset for abdominal
  organ segmentation in surgical data science}.
\newblock \bibinfo{journal}{Scientific Data} \bibinfo{volume}{10},
  \bibinfo{pages}{3}.
\bibitem[{Cheema et~al.(2019)Cheema, Nazir, Sheng, Li, Qin, Kim and
  Feng}]{cheema2019}
\bibinfo{author}{Cheema, M.N.}, \bibinfo{author}{Nazir, A.},
  \bibinfo{author}{Sheng, B.}, \bibinfo{author}{Li, P.}, \bibinfo{author}{Qin,
  J.}, \bibinfo{author}{Kim, J.}, \bibinfo{author}{Feng, D.D.},
  \bibinfo{year}{2019}.
\newblock \bibinfo{title}{Image-aligned dynamic liver reconstruction using
  intra-operative field of views for minimal invasive surgery}.
\newblock \bibinfo{journal}{IEEE Transactions on Biomedical Engineering}
  \bibinfo{volume}{66}, \bibinfo{pages}{2163--2173}.
\bibitem[{Chen et~al.(2017)Chen, Papandreou, Schroff and
  Adam}]{chen2017rethinking}
\bibinfo{author}{Chen, L.C.}, \bibinfo{author}{Papandreou, G.},
  \bibinfo{author}{Schroff, F.}, \bibinfo{author}{Adam, H.},
  \bibinfo{year}{2017}.
\newblock \bibinfo{title}{Rethinking atrous convolution for semantic image
  segmentation}.
\newblock \bibinfo{journal}{arXiv preprint arXiv:1706.05587} .
\bibitem[{Espinel et~al.(2022)Espinel, Calvet, Botros, Buc, Tilmant and
  Bartoli}]{espinel2022}
\bibinfo{author}{Espinel, Y.}, \bibinfo{author}{Calvet, L.},
  \bibinfo{author}{Botros, K.}, \bibinfo{author}{Buc, E.},
  \bibinfo{author}{Tilmant, C.}, \bibinfo{author}{Bartoli, A.},
  \bibinfo{year}{2022}.
\newblock \bibinfo{title}{Using multiple images and contours for deformable
  3d--2d registration of a preoperative ct in laparoscopic liver surgery}.
\newblock \bibinfo{journal}{International Journal of Computer Assisted
  Radiology and Surgery} \bibinfo{volume}{17}, \bibinfo{pages}{2211--2219}.
\bibitem[{Foti et~al.(2020)Foti, Koo, Dowrick, Ramalhinho, Allam, Davidson,
  Stoyanov and Clarkson}]{foti2020intraoperative}
\bibinfo{author}{Foti, S.}, \bibinfo{author}{Koo, B.},
  \bibinfo{author}{Dowrick, T.}, \bibinfo{author}{Ramalhinho, J.},
  \bibinfo{author}{Allam, M.}, \bibinfo{author}{Davidson, B.},
  \bibinfo{author}{Stoyanov, D.}, \bibinfo{author}{Clarkson, M.J.},
  \bibinfo{year}{2020}.
\newblock \bibinfo{title}{Intraoperative liver surface completion with graph
  convolutional vae}, in: \bibinfo{booktitle}{Uncertainty for Safe Utilization
  of Machine Learning in Medical Imaging, and Graphs in Biomedical Image
  Analysis}. \bibinfo{publisher}{Springer}, pp. \bibinfo{pages}{198--207}.
\bibitem[{François et~al.(2020)François, Calvet, Madad~Zadeh, Saboul,
  Gasparini, Samarakoon, Bourdel and Bartoli}]{francois2020}
\bibinfo{author}{François, T.}, \bibinfo{author}{Calvet, L.},
  \bibinfo{author}{Madad~Zadeh, S.}, \bibinfo{author}{Saboul, D.},
  \bibinfo{author}{Gasparini, S.}, \bibinfo{author}{Samarakoon, P.},
  \bibinfo{author}{Bourdel, N.}, \bibinfo{author}{Bartoli, A.},
  \bibinfo{year}{2020}.
\newblock \bibinfo{title}{Detecting the occluding contours of the uterus to
  automatise augmented laparoscopy: Score, loss, dataset, evaluation and
  user-study}.
\newblock \bibinfo{journal}{International Journal of Computer Assisted
  Radiology and Surgery} \bibinfo{volume}{15}.
\bibitem[{Hanocka et~al.(2019)Hanocka, Hertz, Fish, Giryes, Fleishman and
  Cohen-Or}]{hanocka2019meshcnn}
\bibinfo{author}{Hanocka, R.}, \bibinfo{author}{Hertz, A.},
  \bibinfo{author}{Fish, N.}, \bibinfo{author}{Giryes, R.},
  \bibinfo{author}{Fleishman, S.}, \bibinfo{author}{Cohen-Or, D.},
  \bibinfo{year}{2019}.
\newblock \bibinfo{title}{Meshcnn: a network with an edge}.
\newblock \bibinfo{journal}{ACM Transactions on Graphics (TOG)}
  \bibinfo{volume}{38}, \bibinfo{pages}{1--12}.
\bibitem[{Haouchine et~al.(2013)Haouchine, Dequidt, Berger and
  Cotin}]{haouchine2013}
\bibinfo{author}{Haouchine, N.}, \bibinfo{author}{Dequidt, J.},
  \bibinfo{author}{Berger, M.O.}, \bibinfo{author}{Cotin, S.},
  \bibinfo{year}{2013}.
\newblock \bibinfo{title}{Deformation-based augmented reality for hepatic
  surgery}.
\newblock \bibinfo{journal}{Studies in Health Technology and Informatics}
  \bibinfo{volume}{184}, \bibinfo{pages}{182–188}.
\bibitem[{He et~al.(2017)He, Gkioxari, Doll{\'a}r and Girshick}]{he2017mask}
\bibinfo{author}{He, K.}, \bibinfo{author}{Gkioxari, G.},
  \bibinfo{author}{Doll{\'a}r, P.}, \bibinfo{author}{Girshick, R.},
  \bibinfo{year}{2017}.
\newblock \bibinfo{title}{Mask r-cnn}, in: \bibinfo{booktitle}{Proceedings of
  the IEEE international conference on computer vision}, pp.
  \bibinfo{pages}{2961--2969}.
\bibitem[{Isensee et~al.(2021)Isensee, Jaeger, Kohl, Petersen and
  Maier-Hein}]{isensee2021nnu}
\bibinfo{author}{Isensee, F.}, \bibinfo{author}{Jaeger, P.F.},
  \bibinfo{author}{Kohl, S.A.}, \bibinfo{author}{Petersen, J.},
  \bibinfo{author}{Maier-Hein, K.H.}, \bibinfo{year}{2021}.
\newblock \bibinfo{title}{nnu-net: a self-configuring method for deep
  learning-based biomedical image segmentation}.
\newblock \bibinfo{journal}{Nature methods} \bibinfo{volume}{18},
  \bibinfo{pages}{203--211}.
\bibitem[{Jaderberg et~al.(2015)Jaderberg, Simonyan, Zisserman
  et~al.}]{jaderberg2015spatial}
\bibinfo{author}{Jaderberg, M.}, \bibinfo{author}{Simonyan, K.},
  \bibinfo{author}{Zisserman, A.}, et~al., \bibinfo{year}{2015}.
\newblock \bibinfo{title}{Spatial transformer networks}.
\newblock \bibinfo{journal}{Advances in neural information processing systems}
  \bibinfo{volume}{28}.
\bibitem[{Kingma and Ba(2015)}]{kingma2014}
\bibinfo{author}{Kingma, D.}, \bibinfo{author}{Ba, J.}, \bibinfo{year}{2015}.
\newblock \bibinfo{title}{Adam: A method for stochastic optimization}, in:
  \bibinfo{booktitle}{International Conference on Learning Representations
  (ICLR)}, \bibinfo{address}{San Diega, CA, USA}.
\bibitem[{Kipf and Welling(2016)}]{kipf2016semi}
\bibinfo{author}{Kipf, T.N.}, \bibinfo{author}{Welling, M.},
  \bibinfo{year}{2016}.
\newblock \bibinfo{title}{Semi-supervised classification with graph
  convolutional networks}.
\newblock \bibinfo{journal}{arXiv preprint arXiv:1609.02907} .
\bibitem[{Koo et~al.(2017)Koo, {\"O}zg{\"u}r, Le~Roy, Buc and
  Bartoli}]{koo2017}
\bibinfo{author}{Koo, B.}, \bibinfo{author}{{\"O}zg{\"u}r, E.},
  \bibinfo{author}{Le~Roy, B.}, \bibinfo{author}{Buc, E.},
  \bibinfo{author}{Bartoli, A.}, \bibinfo{year}{2017}.
\newblock \bibinfo{title}{Deformable registration of a preoperative 3d liver
  volume to a laparoscopy image using contour and shading cues}, in:
  \bibinfo{booktitle}{Medical Image Computing and Computer Assisted
  Intervention (MICCAI 2017)}, pp. \bibinfo{pages}{326--334}.
\bibitem[{Koo et~al.(2022)Koo, Robu, Allam, Pfeiffer, Thompson, Gurusamy,
  Davidson, Speidel, Hawkes, Stoyanov and Clarkson}]{koo2022}
\bibinfo{author}{Koo, B.}, \bibinfo{author}{Robu, M.R.},
  \bibinfo{author}{Allam, M.}, \bibinfo{author}{Pfeiffer, M.},
  \bibinfo{author}{Thompson, S.}, \bibinfo{author}{Gurusamy, K.},
  \bibinfo{author}{Davidson, B.}, \bibinfo{author}{Speidel, S.},
  \bibinfo{author}{Hawkes, D.}, \bibinfo{author}{Stoyanov, D.},
  \bibinfo{author}{Clarkson, M.J.}, \bibinfo{year}{2022}.
\newblock \bibinfo{title}{Automatic, global registration in laparoscopic liver
  surgery}.
\newblock \bibinfo{journal}{International Journal of Computer Assisted
  Radiology and Surgery} \bibinfo{volume}{17}, \bibinfo{pages}{167--176}.
\bibitem[{Labrunie et~al.(2022)Labrunie, Ribeiro, Mourthadhoi, Tilmant, Le~Roy,
  Buc and Bartoli}]{labrunie2022}
\bibinfo{author}{Labrunie, M.}, \bibinfo{author}{Ribeiro, M.},
  \bibinfo{author}{Mourthadhoi, F.}, \bibinfo{author}{Tilmant, C.},
  \bibinfo{author}{Le~Roy, B.}, \bibinfo{author}{Buc, E.},
  \bibinfo{author}{Bartoli, A.}, \bibinfo{year}{2022}.
\newblock \bibinfo{title}{Automatic preoperative 3d model registration in
  laparoscopic liver resection}.
\newblock \bibinfo{journal}{International Journal of Computer Assisted
  Radiology and Surgery} \bibinfo{volume}{17}, \bibinfo{pages}{1429--1436}.
\bibitem[{Lin et~al.(2014)Lin, Maire, Belongie, Hays, Perona, Ramanan,
  Doll{\'a}r and Zitnick}]{lin2014microsoft}
\bibinfo{author}{Lin, T.Y.}, \bibinfo{author}{Maire, M.},
  \bibinfo{author}{Belongie, S.}, \bibinfo{author}{Hays, J.},
  \bibinfo{author}{Perona, P.}, \bibinfo{author}{Ramanan, D.},
  \bibinfo{author}{Doll{\'a}r, P.}, \bibinfo{author}{Zitnick, C.L.},
  \bibinfo{year}{2014}.
\newblock \bibinfo{title}{{Microsoft COCO: Common objects in context}}, in:
  \bibinfo{booktitle}{European conference on computer vision}, pp.
  \bibinfo{pages}{740--755}.
\bibitem[{Liu et~al.(2019)Liu, Li, Chen and Li}]{liu2019soft}
\bibinfo{author}{Liu, S.}, \bibinfo{author}{Li, T.}, \bibinfo{author}{Chen,
  W.}, \bibinfo{author}{Li, H.}, \bibinfo{year}{2019}.
\newblock \bibinfo{title}{Soft rasterizer: A differentiable renderer for
  image-based 3d reasoning}, in: \bibinfo{booktitle}{Proceedings of the
  IEEE/CVF International Conference on Computer Vision}, pp.
  \bibinfo{pages}{7708--7717}.
\bibitem[{Luo et~al.(2020)Luo, Yin, Zhang, Xiao, He, Meng, Zhang, Cai, He,
  Zhang, Hu, Guo, Liang, Zhou, Liu, Sun, Guo, Fang, Liu and Jia}]{luo2019}
\bibinfo{author}{Luo, H.}, \bibinfo{author}{Yin, D.}, \bibinfo{author}{Zhang,
  S.}, \bibinfo{author}{Xiao, D.}, \bibinfo{author}{He, B.},
  \bibinfo{author}{Meng, F.}, \bibinfo{author}{Zhang, Y.},
  \bibinfo{author}{Cai, W.}, \bibinfo{author}{He, S.}, \bibinfo{author}{Zhang,
  W.}, \bibinfo{author}{Hu, Q.}, \bibinfo{author}{Guo, H.},
  \bibinfo{author}{Liang, S.}, \bibinfo{author}{Zhou, S.},
  \bibinfo{author}{Liu, S.}, \bibinfo{author}{Sun, L.}, \bibinfo{author}{Guo,
  X.}, \bibinfo{author}{Fang, C.}, \bibinfo{author}{Liu, L.},
  \bibinfo{author}{Jia, F.}, \bibinfo{year}{2020}.
\newblock \bibinfo{title}{Augmented reality navigation for liver resection with
  a stereoscopic laparoscope}.
\newblock \bibinfo{journal}{Computer Methods and Programs in Biomedicine}
  \bibinfo{volume}{187}.
\bibitem[{Modrzejewski et~al.(2019)Modrzejewski, Collins, Seeliger, Bartoli,
  Hostettler and Marescaux}]{modrzejewski2019}
\bibinfo{author}{Modrzejewski, R.}, \bibinfo{author}{Collins, T.},
  \bibinfo{author}{Seeliger, B.}, \bibinfo{author}{Bartoli, A.},
  \bibinfo{author}{Hostettler, A.}, \bibinfo{author}{Marescaux, J.},
  \bibinfo{year}{2019}.
\newblock \bibinfo{title}{An in vivo porcine dataset and evaluation methodology
  to measure soft-body laparoscopic liver registration accuracy with an
  extended algorithm that handles collisions}.
\newblock \bibinfo{journal}{International Journal of Computer Assisted
  Radiology and Surgery} \bibinfo{volume}{14}, \bibinfo{pages}{1237--1245}.
\bibitem[{Müller et~al.(2022)Müller, Soto-Rey and Kramer}]{muller2022}
\bibinfo{author}{Müller, D.}, \bibinfo{author}{Soto-Rey, I.},
  \bibinfo{author}{Kramer, F.}, \bibinfo{year}{2022}.
\newblock \bibinfo{title}{Towards a guideline for evaluation metrics in medical
  image segmentation}.
\newblock \bibinfo{journal}{BMC Research Notes} \bibinfo{volume}{15}.
\newblock \DOIprefix\doi{10.1186/s13104-022-06096-y}.
\bibitem[{Oktay et~al.(2018)Oktay, Schlemper, Folgoc, Lee, Heinrich, Misawa,
  Mori, McDonagh, Hammerla, Kainz, Glocker and Rueckert}]{oktay2018}
\bibinfo{author}{Oktay, O.}, \bibinfo{author}{Schlemper, J.},
  \bibinfo{author}{Folgoc, L.L.}, \bibinfo{author}{Lee, M.},
  \bibinfo{author}{Heinrich, M.}, \bibinfo{author}{Misawa, K.},
  \bibinfo{author}{Mori, K.}, \bibinfo{author}{McDonagh, S.},
  \bibinfo{author}{Hammerla, N.Y.}, \bibinfo{author}{Kainz, B.},
  \bibinfo{author}{Glocker, B.}, \bibinfo{author}{Rueckert, D.},
  \bibinfo{year}{2018}.
\newblock \bibinfo{title}{{Attention U-Net: Learning Where to Look for the
  Pancreas}}, in: \bibinfo{booktitle}{Medical Imaging with Deep Learning}.
\bibitem[{Qi et~al.(2017)Qi, Yi, Su and Guibas}]{qi2017pointnet++}
\bibinfo{author}{Qi, C.R.}, \bibinfo{author}{Yi, L.}, \bibinfo{author}{Su, H.},
  \bibinfo{author}{Guibas, L.J.}, \bibinfo{year}{2017}.
\newblock \bibinfo{title}{Pointnet++: Deep hierarchical feature learning on
  point sets in a metric space}.
\newblock \bibinfo{journal}{Advances in neural information processing systems}
  \bibinfo{volume}{30}.
\bibitem[{{Radboud University Medical Center}(2023)}]{grandchallenge}
\bibinfo{author}{{Radboud University Medical Center}}, \bibinfo{year}{2023}.
\newblock \bibinfo{title}{{Grand Challenge}}.
\newblock \URLprefix \url{https://www.grand-challenge.org/}.
\bibitem[{Ravi et~al.(2020)Ravi, Reizenstein, Novotny, Gordon, Lo, Johnson and
  Gkioxari}]{ravi2020accelerating}
\bibinfo{author}{Ravi, N.}, \bibinfo{author}{Reizenstein, J.},
  \bibinfo{author}{Novotny, D.}, \bibinfo{author}{Gordon, T.},
  \bibinfo{author}{Lo, W.Y.}, \bibinfo{author}{Johnson, J.},
  \bibinfo{author}{Gkioxari, G.}, \bibinfo{year}{2020}.
\newblock \bibinfo{title}{Accelerating 3d deep learning with pytorch3d}.
\newblock \bibinfo{journal}{arXiv preprint arXiv:2007.08501} .
\bibitem[{Redmon et~al.(2016)Redmon, Divvala, Girshick and
  Farhadi}]{redmon2016you}
\bibinfo{author}{Redmon, J.}, \bibinfo{author}{Divvala, S.},
  \bibinfo{author}{Girshick, R.}, \bibinfo{author}{Farhadi, A.},
  \bibinfo{year}{2016}.
\newblock \bibinfo{title}{You only look once: Unified, real-time object
  detection}, in: \bibinfo{booktitle}{Proceedings of the IEEE conference on
  computer vision and pattern recognition}, pp. \bibinfo{pages}{779--788}.
\bibitem[{Robu et~al.(2018)Robu, Ramalhinho, Thompson, Gurusamy, Davidson,
  Hawkes, Stoyanov and Clarkson}]{robu2018}
\bibinfo{author}{Robu, M.R.}, \bibinfo{author}{Ramalhinho, J.},
  \bibinfo{author}{Thompson, S.}, \bibinfo{author}{Gurusamy, K.},
  \bibinfo{author}{Davidson, B.}, \bibinfo{author}{Hawkes, D.},
  \bibinfo{author}{Stoyanov, D.}, \bibinfo{author}{Clarkson, M.J.},
  \bibinfo{year}{2018}.
\newblock \bibinfo{title}{Global rigid registration of ct to video in
  laparoscopic liver surgery}.
\newblock \bibinfo{journal}{International Journal of Computer Assisted
  Radiology and Surgery} \bibinfo{volume}{13}, \bibinfo{pages}{947--956}.
\bibitem[{Ronneberger et~al.(2015)Ronneberger, Fischer and
  Brox}]{ronneberger2015u}
\bibinfo{author}{Ronneberger, O.}, \bibinfo{author}{Fischer, P.},
  \bibinfo{author}{Brox, T.}, \bibinfo{year}{2015}.
\newblock \bibinfo{title}{U-net: Convolutional networks for biomedical image
  segmentation}, in: \bibinfo{booktitle}{International Conference on Medical
  image computing and computer-assisted intervention},
  \bibinfo{organization}{Springer}. pp. \bibinfo{pages}{234--241}.
\bibitem[{Soler et~al.(2014)Soler, Nicolau, Pessaux, Mutter and
  Marescaux}]{soler2014}
\bibinfo{author}{Soler, L.}, \bibinfo{author}{Nicolau, S.},
  \bibinfo{author}{Pessaux, P.}, \bibinfo{author}{Mutter, D.},
  \bibinfo{author}{Marescaux, J.}, \bibinfo{year}{2014}.
\newblock \bibinfo{title}{Real-time 3d image reconstruction guidance in liver
  resection surgery}.
\newblock \bibinfo{journal}{Hepatobiliary Surgery and Nutrition}
  \bibinfo{volume}{3}.
\bibitem[{Taha et~al.(2014)Taha, Hanbury and del Toro}]{taha2014}
\bibinfo{author}{Taha, A.A.}, \bibinfo{author}{Hanbury, A.},
  \bibinfo{author}{del Toro, O.A.J.}, \bibinfo{year}{2014}.
\newblock \bibinfo{title}{A formal method for selecting evaluation metrics for
  image segmentation}, in: \bibinfo{booktitle}{2014 IEEE International
  Conference on Image Processing (ICIP)}, pp. \bibinfo{pages}{932--936}.
\bibitem[{Thompson et~al.(2015)Thompson, Totz, Song, Johnsen, Stoyanov,
  Ourselin, Gurusamy, Schneider, Davidson, Hawkes and Clarkson}]{thompson2015}
\bibinfo{author}{Thompson, S.}, \bibinfo{author}{Totz, J.},
  \bibinfo{author}{Song, Y.}, \bibinfo{author}{Johnsen, S.},
  \bibinfo{author}{Stoyanov, D.}, \bibinfo{author}{Ourselin, S.},
  \bibinfo{author}{Gurusamy, K.}, \bibinfo{author}{Schneider, C.},
  \bibinfo{author}{Davidson, B.}, \bibinfo{author}{Hawkes, D.},
  \bibinfo{author}{Clarkson, M.J.}, \bibinfo{year}{2015}.
\newblock \bibinfo{title}{{Accuracy validation of an image guided laparoscopy
  system for liver resection}}, in: \bibinfo{booktitle}{Medical Imaging 2015:
  Image-Guided Procedures, Robotic Interventions, and Modeling},
  \bibinfo{publisher}{SPIE}. pp. \bibinfo{pages}{52 -- 63}.
\bibitem[{Twinanda et~al.(2017)Twinanda, Shehata, Mutter, Marescaux,
  de~Mathelin and Padoy}]{twinanda2016}
\bibinfo{author}{Twinanda, A.P.}, \bibinfo{author}{Shehata, S.},
  \bibinfo{author}{Mutter, D.}, \bibinfo{author}{Marescaux, J.},
  \bibinfo{author}{de~Mathelin, M.}, \bibinfo{author}{Padoy, N.},
  \bibinfo{year}{2017}.
\newblock \bibinfo{title}{{EndoNet: A} deep architecture for recognition tasks
  on laparoscopic videos}.
\newblock \bibinfo{journal}{IEEE transactions on medical imaging}
  \bibinfo{volume}{36}, \bibinfo{pages}{86—97}.
\newblock \DOIprefix\doi{10.1109/tmi.2016.2593957}.
\bibitem[{{Unity Technologies}(2023)}]{unity_store}
\bibinfo{author}{{Unity Technologies}}, \bibinfo{year}{2023}.
\newblock \bibinfo{title}{{Unity Asset Store}}.
\newblock \URLprefix \url{https://assetstore.unity.com/}.
\bibitem[{Wagner et~al.(2023)Wagner, Müller-Stich, Kisilenko, Tran, Heger,
  Mündermann, Lubotsky, Müller, Davitashvili, Capek, Reinke, Reid, Yu,
  Vardazaryan, Nwoye, Padoy, Liu, Lee, Disch, Meine, Xia, Jia, Kondo, Reiter,
  Jin, Long, Jiang, Dou, Heng, Twick, Kirtac, Hosgor, Bolmgren, Stenzel, {von
  Siemens}, Zhao, Ge, Sun, Xie, Guo, Liu, Kenngott, Nickel, von Frankenberg,
  Mathis-Ullrich, Kopp-Schneider, Maier-Hein, Speidel and
  Bodenstedt}]{wagner2023}
\bibinfo{author}{Wagner, M.}, \bibinfo{author}{Müller-Stich, B.P.},
  \bibinfo{author}{Kisilenko, A.}, \bibinfo{author}{Tran, D.},
  \bibinfo{author}{Heger, P.}, \bibinfo{author}{Mündermann, L.},
  \bibinfo{author}{Lubotsky, D.M.}, \bibinfo{author}{Müller, B.},
  \bibinfo{author}{Davitashvili, T.}, \bibinfo{author}{Capek, M.},
  \bibinfo{author}{Reinke, A.}, \bibinfo{author}{Reid, C.},
  \bibinfo{author}{Yu, T.}, \bibinfo{author}{Vardazaryan, A.},
  \bibinfo{author}{Nwoye, C.I.}, \bibinfo{author}{Padoy, N.},
  \bibinfo{author}{Liu, X.}, \bibinfo{author}{Lee, E.J.},
  \bibinfo{author}{Disch, C.}, \bibinfo{author}{Meine, H.},
  \bibinfo{author}{Xia, T.}, \bibinfo{author}{Jia, F.}, \bibinfo{author}{Kondo,
  S.}, \bibinfo{author}{Reiter, W.}, \bibinfo{author}{Jin, Y.},
  \bibinfo{author}{Long, Y.}, \bibinfo{author}{Jiang, M.},
  \bibinfo{author}{Dou, Q.}, \bibinfo{author}{Heng, P.A.},
  \bibinfo{author}{Twick, I.}, \bibinfo{author}{Kirtac, K.},
  \bibinfo{author}{Hosgor, E.}, \bibinfo{author}{Bolmgren, J.L.},
  \bibinfo{author}{Stenzel, M.}, \bibinfo{author}{{von Siemens}, B.},
  \bibinfo{author}{Zhao, L.}, \bibinfo{author}{Ge, Z.}, \bibinfo{author}{Sun,
  H.}, \bibinfo{author}{Xie, D.}, \bibinfo{author}{Guo, M.},
  \bibinfo{author}{Liu, D.}, \bibinfo{author}{Kenngott, H.G.},
  \bibinfo{author}{Nickel, F.}, \bibinfo{author}{von Frankenberg, M.},
  \bibinfo{author}{Mathis-Ullrich, F.}, \bibinfo{author}{Kopp-Schneider, A.},
  \bibinfo{author}{Maier-Hein, L.}, \bibinfo{author}{Speidel, S.},
  \bibinfo{author}{Bodenstedt, S.}, \bibinfo{year}{2023}.
\newblock \bibinfo{title}{Comparative validation of machine learning algorithms
  for surgical workflow and skill analysis with the heichole benchmark}.
\newblock \bibinfo{journal}{Medical Image Analysis} \bibinfo{volume}{86},
  \bibinfo{pages}{102770}.
\bibitem[{Wu et~al.(2021)Wu, Pan, Zhang, WANG, Liu and Lin}]{wu2021}
\bibinfo{author}{Wu, T.}, \bibinfo{author}{Pan, L.}, \bibinfo{author}{Zhang,
  J.}, \bibinfo{author}{WANG, T.}, \bibinfo{author}{Liu, Z.},
  \bibinfo{author}{Lin, D.}, \bibinfo{year}{2021}.
\newblock \bibinfo{title}{Balanced chamfer distance as a comprehensive metric
  for point cloud completion}, in: \bibinfo{editor}{Ranzato, M.},
  \bibinfo{editor}{Beygelzimer, A.}, \bibinfo{editor}{Dauphin, Y.},
  \bibinfo{editor}{Liang, P.}, \bibinfo{editor}{Vaughan, J.W.} (Eds.),
  \bibinfo{booktitle}{Advances in Neural Information Processing Systems},
  \bibinfo{publisher}{Curran Associates, Inc.}. pp.
  \bibinfo{pages}{29088--29100}.
\newblock \URLprefix
  \url{https://proceedings.neurips.cc/paper_files/paper/2021/file/f3bd5ad57c8389a8a1a541a76be463bf-Paper.pdf}.
\bibitem[{Yang and Soatto(2020)}]{yang2020fda}
\bibinfo{author}{Yang, Y.}, \bibinfo{author}{Soatto, S.}, \bibinfo{year}{2020}.
\newblock \bibinfo{title}{Fda: Fourier domain adaptation for semantic
  segmentation}, in: \bibinfo{booktitle}{Proceedings of the IEEE/CVF Conference
  on Computer Vision and Pattern Recognition}, pp. \bibinfo{pages}{4085--4095}.
\bibitem[{Zhang et~al.(2018)Zhang, Liu and Wang}]{zhang2018road}
\bibinfo{author}{Zhang, Z.}, \bibinfo{author}{Liu, Q.}, \bibinfo{author}{Wang,
  Y.}, \bibinfo{year}{2018}.
\newblock \bibinfo{title}{Road extraction by deep residual u-net}.
\newblock \bibinfo{journal}{IEEE Geoscience and Remote Sensing Letters}
  \bibinfo{volume}{15}, \bibinfo{pages}{749--753}.
\bibitem[{Zhou et~al.(2018)Zhou, Rahman~Siddiquee, Tajbakhsh and
  Liang}]{zhou2018unet++}
\bibinfo{author}{Zhou, Z.}, \bibinfo{author}{Rahman~Siddiquee, M.M.},
  \bibinfo{author}{Tajbakhsh, N.}, \bibinfo{author}{Liang, J.},
  \bibinfo{year}{2018}.
\newblock \bibinfo{title}{Unet++: A nested u-net architecture for medical image
  segmentation}, in: \bibinfo{booktitle}{Deep learning in medical image
  analysis and multimodal learning for clinical decision support}.
  \bibinfo{publisher}{Springer}, pp. \bibinfo{pages}{3--11}.

\end{thebibliography}
\end{document}